\documentclass[10pt,journal,compsoc]{IEEEtran}
\ifCLASSOPTIONcompsoc
  \usepackage[nocompress]{cite}
\else
  \usepackage{cite}
\fi
\ifCLASSINFOpdf
\else
\fi
\usepackage{amsmath,bm}
\usepackage{amsfonts,amssymb}
\usepackage{graphicx}
\usepackage{caption}
\usepackage{subfigure}

\usepackage{subfloat}

\usepackage{booktabs}
\usepackage{multirow}
\usepackage{color}
\usepackage{algorithm}
\usepackage{algorithmic}
\usepackage[switch]{{lineno}}
\usepackage{colortbl}
\definecolor{mygray}{gray}{.9}
\usepackage{dsfont}
\usepackage{bbding}
\usepackage{pifont}
\usepackage{wasysym}
\usepackage{amssymb}

\newcommand{\xmark}{\ding{55}}

\label{definition}

\def\bx{{\bm{x}}}

\def\bz{{\bm{z}}}

\begin{document}
%
\title{Scale-Invariant Adversarial Attack for Evaluating and Enhancing Adversarial Defenses}
%
%
%
%

\author{Mengting Xu,
        Tao Zhang, Zhongnian Li, 
        and Daoqiang Zhang
\IEEEcompsocitemizethanks{\IEEEcompsocthanksitem Mengting Xu, Tao Zhang, Zhongnian Li, and Daoqiang Zhang are with the College of Computer Science and Technology, Nanjing University of Aeronautics and Astronautics, Nanjing 211106, China.\protect\\
E-mail: xumengting1203@163.com, lrhselous@nuaa.edu.cn, zhongnianli@163.com, dqzhang@nuaa.edu.cn}
}

%
%

\markboth{Journal of \LaTeX\ Class Files,~Vol.~20, No.~11, August~2021}%
{Shell \MakeLowercase{\textit{et al.}}: Bare Demo of IEEEtran.cls for Computer Society Journals}
%



\IEEEtitleabstractindextext{%
\begin{abstract}
Efficient and effective attacks are crucial for reliable evaluation of defenses, and also for developing robust models. Projected Gradient Descent (PGD) attack has been demonstrated to be one of the most successful adversarial attacks. However, the effect of the standard PGD attack can be easily weakened by rescaling the logits, while the original decision of every input will not be changed.
To mitigate this issue, in this paper, we propose Scale-Invariant Adversarial Attack (SI-PGD), which utilizes the angle between the features in the penultimate layer and the weights in the softmax layer to guide the generation of adversaries. 
The cosine angle matrix is used to learn angularly discriminative representation and will not be changed with the rescaling of logits, thus making SI-PGD attack to be stable and effective.
We evaluate our attack against multiple defenses and show improved performance when compared with existing attacks. 
Further, we propose Scale-Invariant (SI) adversarial defense mechanism based on the cosine angle matrix, which can be embedded into the popular adversarial defenses.
The experimental results show the defense method with our SI mechanism achieves state-of-the-art performance among multi-step and single-step defenses.
\end{abstract}

}

\maketitle

\IEEEdisplaynontitleabstractindextext

%
\IEEEpeerreviewmaketitle

\IEEEraisesectionheading{\section{Introduction}\label{sec:introduction}}

%
%
%
%

\IEEEPARstart{D}{espite} the widespread success of neural networks, recent studies have highlighted the lack of robustness in state-of-the-art neural network models, e.g., a well-trained network can be easily misled by a visually imperceptible adversarial image.
The vulnerability to adversarial examples calls into question the safety-critical applications and services deployed by neural networks. 
This finding has also spurred immense interest towards defense methods to improve the robustness of deep neural networks against adversarial attacks. While initial attempts of improving robustness against adversarial attacks used just single-step adversaries for training~\cite{goodfellow2014explaining}, they were later shown to be ineffective against strong multi-step attacks~\cite{kurakin2016adversarial}.
Then the defenses using randomised or non-differentiable components were proposed to minimise the effectiveness of  gradient-based adversarial attack. However, many such defenses~\cite{buckman2018thermometer,xie2018mitigating,guo2018countering} were later broken by Athalye et al.~\cite{athalye2018obfuscated}.
This game of constructing defense methods against existing attacks and constructing stronger attacks against the proposed defenses is crucial to the progress of this field.
Therefore, it's necessary to construct effective adversarial attack for evaluating and enhancing adversarial defense.

\begin{figure}[t]
	\centering                                                      
	\includegraphics[width=0.9\columnwidth]{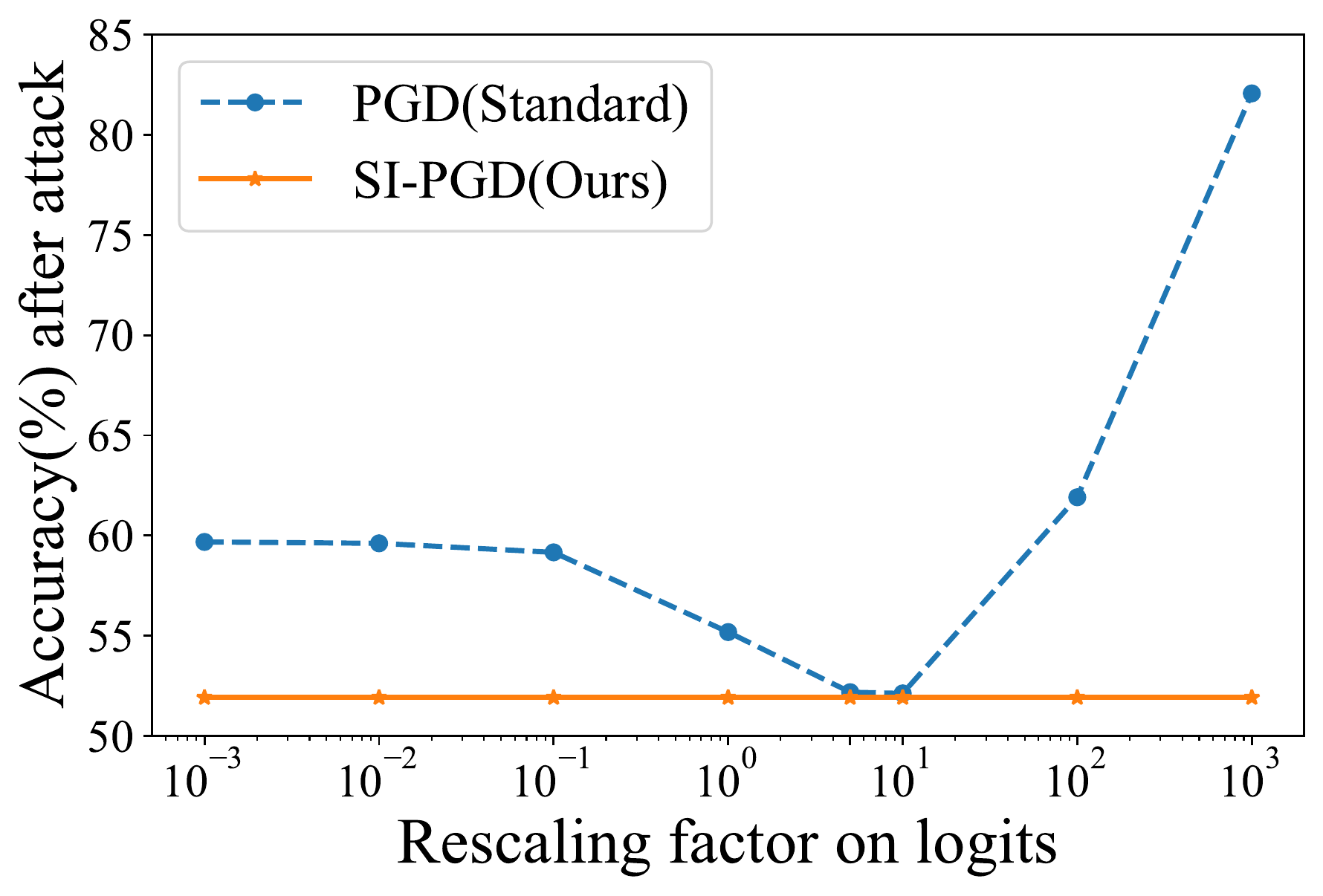}  
	\caption{The unstable performance of PGD attack. The base model is trained by TRADES~\cite{zhang2019theoretically}. When the logits of the model rescaled by factor from $10^{-3}$ to $10^3$, the model accuracy attacked by PGD (bule line) changes obviously, while our Scale-Invariant adversarial attack (SI-PGD,yellow line) is stable and effective.}   
	\label{fig:motivation} 
\end{figure}

One of the most effective adversarial attacks to date is Projected Gradient Descent (PGD) attack~\cite{madry2018towards}, which uniformly samples the initial point from the neighborhood around the clean input, and computes the gradient of the cross-entropy (CE) loss function around the current output logits of the model to update the perturbations.
However, despite the high performance of the PGD attack, it still has a fatal flaw. The effect of the PGD attack with CE loss based on logits can be easily weakened by rescaling the logits while the original decision of the each input will not be changed. As shown in Fig.~\ref{fig:motivation}, when the rescaling factor mutiplied to logits changes from $10^{-3}$ to $10^3$, the effect of PGD attack has also changed significantly. The instability of PGD attack may make the robustness of model evaluated by PGD attack become unconvincing~\cite{atzmon2019controlling,croce2020reliable}. Even worse, the performance of those PGD-based defense models may also be weakened.

To address this problem, in this paper, we propose the Scale-Invariant adversarial attack (SI-PGD) by utilizing the angle between the features in the penultimate layer and the weights in the softmax layer with CE loss to guide the generation of adversarial examples. Specifically, the cosine angle matrix is used to learn angularly discriminative representation and will not be changed with the rescaling of logits, thereby making SI-PGD effective and stable. We demonstrate the state-of-the-art attack efficacy on multiple defenses and datasets of the proposed SI-PGD. 
We further propose the Scale-Invariant (SI) adversarial defense mechanism based on the cosine angle matrix, which is concise and adaptable to be embedded into the popular adversarial defenses.
With proposed SI mechanism, we observe significant improvements in both multi-step and single-step adversarial defenses.

Our main contributions are:
\begin{itemize}
	\item We reveal the fact that the effect of PGD attack can be easily changed with the rescaling of the logits in Fig.~\ref{fig:motivation}.
	\item We propose Scale-Invariant adversarial attack (SI-PGD), which achieves the state-of-the-art performance across multiple defenses for a single attack and multiple random restarts.
	\item We propose Scale-Invariant (SI) adversarial defense mechanism, which achieves the state-of-the-art performance of the model robustness when embedded into the existing multi-step and single-step adversarial defenses.
\end{itemize}

The remainder of the paper is organized as follows. In
Section 2 we review related recent literature. We describe the materials including the datasets and the problem formulations in Section 3. 
In Section 4, we reveal the shortcoming of PGD attack, and propose our Scale-Invariant adversarial attack (SI-PGD) and Scale-Invariant (SI) adversarial defense mechanism. In Section 5, we evaluate the effectiveness of SI-PGD attack and SI defense mechanism.
We conclude in Section 6.
 

\section{Related Works}
\subsection{Adversarial attacks}
A lot of adversarial attacks have been proposed in recent years. Fast Gradient Sign Method (FGSM)~\cite{goodfellow2014explaining} is one of the earliest attacks specific to $L_\infty $ constrained adversaries. It calculates the gradient of the loss function with respect to the pixel, and modifies the pixel value of a fixed step along the direction of the gradient. Kurakin et al.~\cite{kurakin2016adversarial} introduced a significantly stronger, multi-step variant of this attack called Iterative FGSM (I-FGSM). Then Madry et al.~\cite{madry2018towards} developed a variant of this attack, which involves the addition of initial random noise to the clean image, and is commonly referred as Projected Gradient Descent (PGD) attack.
Carlini and Wagner~\cite{carlini2017towards} explored the use of maximum margin loss and optimization methods to craft adversarial examples with high fidelity and small distortion with respect to the original image.
More recently, Croce and Hein~\cite{croce2020reliable} proposed AutoPGD, which is an automatised variant of the PGD attack, that uses a step-learning rate schedule adaptively based on the past progression of the optimization. They further introduced a new loss function, Difference of Logits Ratio (DLR), which is a scale invariant version of the maximum margin loss on logits. Additionally, they proposed AutoAttack (AA), an ensemble of AutoPGD with the CE loss and the DLR loss, the FAB attack~\cite{croce2020minimally} and Square attack~\cite{andriushchenko2020square}.

\subsection{Defenses against adversarial attacks}

One of the most successful empirical defenses to date is adversarial training (AT). It was first proposed in~\cite{szegedy2014intriguing,goodfellow2014explaining}, where they showed that adding adversarial examples to the training set can improve the model robustness against attacks. More recently, Madry et al.~\cite{madry2018towards} formulated adversarial training as a min-max optimization problem and demonstrated that adversarial training with PGD attack leads to empirical robust model. 
Following this, Kannan et al.~\cite{kannan2018adversarial} proposed adversarial logits paring (ALP), which regularizes the distance between the clean logits and the adversarial ones.
Then Zhang et al.~\cite{zhang2019theoretically} presented a tight upper bound on the gap between natural and robust error, in order to quantify the trade-off between accuracy and robustness. Using the theory of classification calibrated losses, they developed TRADES, a multi-step gradient-based technique.
Wang et al.~\cite{wang2019improving} further proposed MART which differentiates the misclassified examples and correctly classified examples during adversarial training, and adopted a regularized adversarial loss involving both adversarial and natural examples to improve the robustness of models. 
However, methods such as AT, TRADES are computationally intensive, as they inherently depend upon the generation of strong adversaries through iterative attacks.

Consequently, efforts were made to develop techniques that accelerated adversarial training. Shafahi et al.~\cite{shafahi2019adversarial} proposed a variant of AT, known as Adversarial Training for Free (FreeAT), where the gradients accumulated in each step are used to simultaneously update the adversarial example as well as the network parameters.
Contrary to prior wisdom, Wong et al.~\cite{wong2019fast} found the surprising result that Randomised FGSM (R-FGSM)~\cite{tramer2018ensemble} training could indeed be successfully utilized to produce robust models. It was shown that R-FGSM adversarial training (RFGSM-AT) could be made effective with the use of small-step sizes for generation of adversarial examples, in combination with other techniques such as early-stopping and cyclic learning rates. 
Moreover, Sriramanan et al.~\cite{sriramanan2020guided} utilized function mapping of the clean image to guide the generation of adversaries for adversarial training (GAT), and Andriushchenko et al.~\cite{andriushchenko2020understanding} proposed GradAlign, that prevents catastrophic overfitting by explicitly maximizing the gradient alignment inside the perturbation set.

\section{Materials}
	
\subsection{Datasets}
We use the CIFAR-10, CIFAR-100~\cite{krizhevsky2009learning}, and SVHN~\cite{netzer2011reading} datasets in our study. The CIFAR-10 dataset consists of 60,000 32x32 colour images in 10 classes, with 6,000 images per class. There are 50,000 training images and 10,000 test images. The CIFAR-100 has 100 classes containing 600 images each. There are 500 training images and 100 testing images per class. The 100 classes in the CIFAR-100 are grouped into 20 superclasses. The SVHN dataset contains colour images of streetview house numbers for 10 different classes.

\subsection{Problem formulations}
Let $\mathcal{X} \in \mathbb{R}^d$ denote the input space and $\mathcal{Y}=\{1,\cdots, K\}$ be a finite set consists of $K$ possible class labels. $D=\{(\bm{x}_1,y_1),\cdots,(\bm{x}_m,y_m)\}$ is a training set with $m$ labeled examples, where $\bm{x}_i\in \mathcal{X}$ is the feature vector and $y_i\in \mathcal{Y}$ is the label of the $i$-th example. 
For the classification task with $K$ labels in $[K]:=\{1,\cdots,K\}$, a deep neural network can be generally denoted as the mapping function $f(\bx)$ for input $\bx$ as
\begin{equation}
\label{equ:f}
f(\bx)=\mathbb{S}(\bm{W}^\top\bm{z}+b),
\end{equation}
where we denote $g(\bx)=\bm{W}^\top\bm{z}+b$ as the output logits of the model, $\bm{z}=\bm{z}(\bx;\bm{w})$ is the extracted feature with model parameters $\bm{w}$, the matrix $\bm{W}=(W_1,\cdots,W_K)$ and vector $b$ are the weight and bias in the softmax layer, respectively, and $\mathbb{S}(\cdot)$ is the softmax function.

Next, we introduce the detailed information of PGD (standard PGD attack with cross-entropy loss based on the output logits of the model~\cite{madry2018towards}), PGDCW (PGD attack with maximum margin loss proposed in C\&W~\cite{carlini2017towards}), and PGDLR (PGD attack with difference of logits ratio loss (DLR)~\cite{croce2020reliable}) attacks we used to compare with our proposed SI-PGD in our study.

\subsubsection{PGD}
The standard PGD attack with cross-entropy loss is based on the output logits of the model~\cite{madry2018towards}. The formulation of PGD can be described as:
\begin{align}
\label{equ:PGD1}
&\bx_0^\prime = \bx+\mathbb{U}(-\epsilon,+\epsilon), \\
\bx_{t+1}^\prime = {\rm Clip}_{\bx,\epsilon}&(\bx_t^\prime +\eta \cdot {\rm sign}(\nabla_\bx \mathcal{L}(f(\bx_t^\prime),y))), \\
\mathcal{L}(f(\bx_t^\prime),y) &= \mathcal{L}_{\rm CE}(f(\bx_t^\prime),y)= -\log f(\bx_t^\prime)_y,
\end{align}
where $\mathbb{U}(\cdot)$ is the uniform function, ${\rm Clip}_{\bx,\epsilon}(\cdot)$ projects the adversarial example to satisfy the $\epsilon$ constrain, $t$ is the number of step and $\eta$ is the step size. $\mathcal{L}_{\rm CE}(\cdot)$ is the cross-entropy loss function.

\subsubsection{PGDCW}
PGDCW is the PGD attack with maximum margin loss proposed in C\&W~\cite{carlini2017towards} rather than standard cross-entropy loss:
\begin{equation}
\mathcal{L}(f(\bx_t^\prime),y) = \mathcal{L}_{\rm CW}(f(\bx_t^\prime),y)= -f(\bx_t^\prime)_y+ \max_{i \neq y} f(\bx_t^\prime)_i.
\end{equation}

\subsubsection{PGDLR}
PGDLR is the PGD attack with difference of logits ratio loss (DLR)~\cite{croce2020reliable}:
\begin{equation}
\begin{split}
\mathcal{L}(f(\bx_t^\prime),y)& = \mathcal{L}_{\rm DLR}(f(\bx_t^\prime),y) \\& = - \frac{f(\bx_t^\prime)_y-\max_{i \neq y} f(\bx_t^\prime)_i}{f(\bx_t^\prime)_{\pi_1}-f(\bx_t^\prime)_{\pi_3}},
\end{split}
\end{equation}
where $\pi$ is the ordering of the components of $f(\bx_t^\prime)$ in decreasing order.

\begin{figure*}[t]
	\centering  
	\subfigure[Misclassified]{
		\label{fig:defense1}
		\begin{minipage}{0.6\columnwidth}
			\centering                                                          
			\includegraphics[width=1\columnwidth]{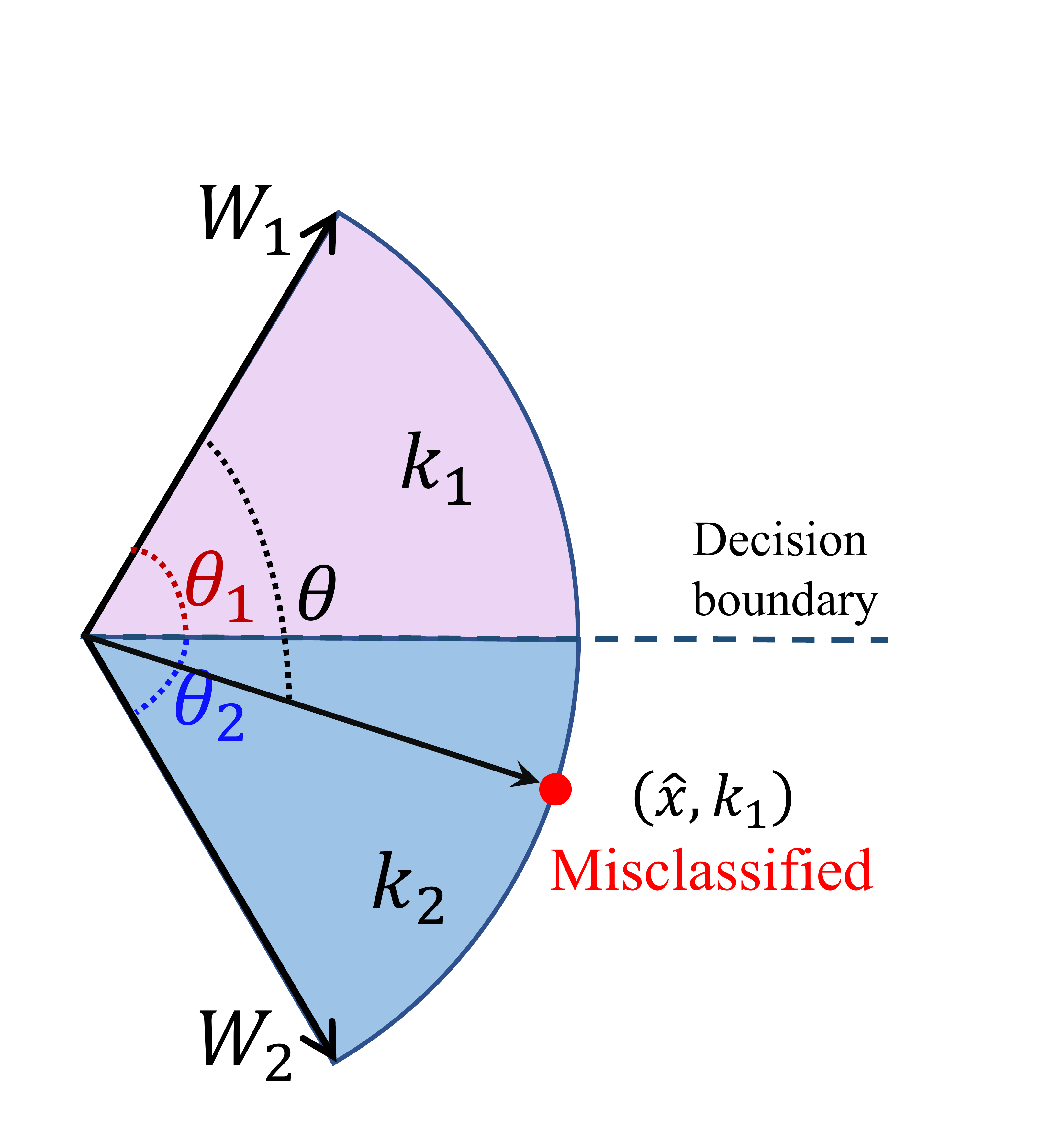}           
	\end{minipage}}
	\subfigure[Updating the norm of weight]{ 
		\label{fig:defense2}                  
		\begin{minipage}{0.6\columnwidth}
			\centering                                                          
			\includegraphics[width=1\columnwidth]{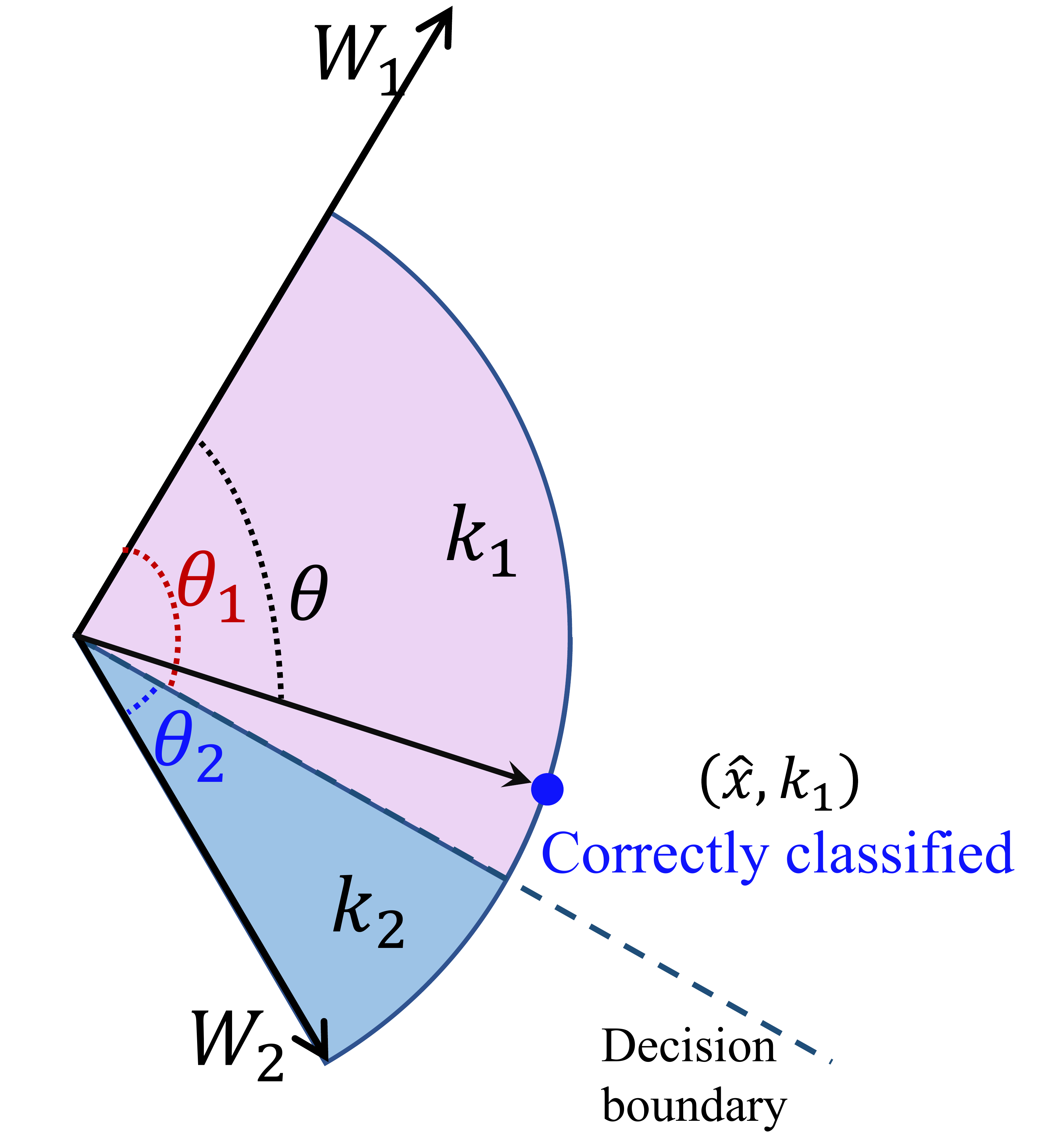}           
	\end{minipage}}
	\subfigure[Updating the direction of the weight]{  
		\label{fig:defense3}                	  
		\begin{minipage}{0.6\columnwidth}
			\centering                                                          
			\includegraphics[width=1\columnwidth]{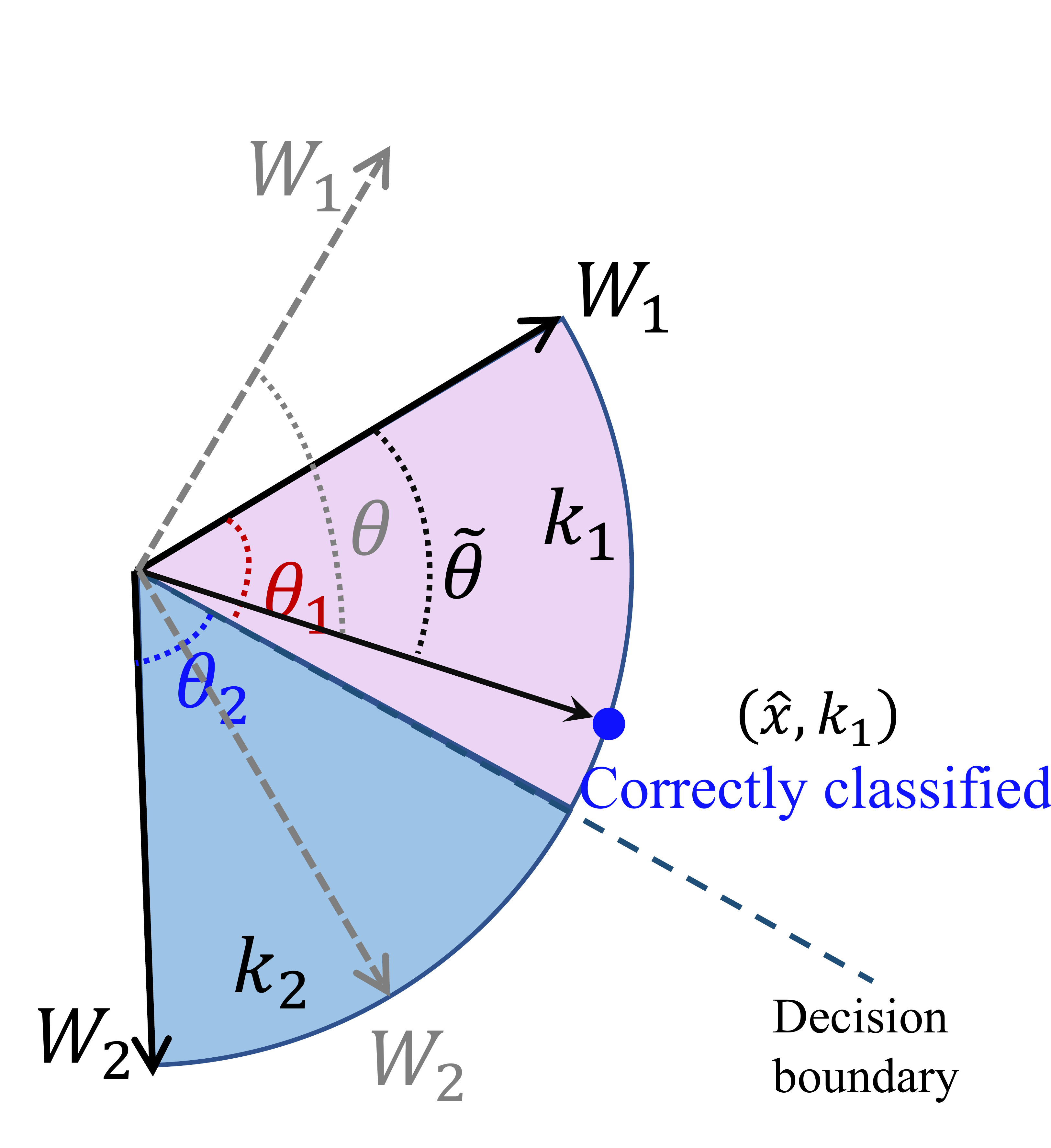}               
	\end{minipage}}     
	\caption{A geometrical interpretation of our proposed SI mechanism. Different color areas represent distinct classes. (a) A adversarial example $\hat{\bx}$ with label $k_1$ is misclassified by the model. (b) The model can correctly classified adversarial example by updating the norm of weight. However, the endless increase in weight may lead the phenomenon of gradient vanishing. (c) The model can correctly classified adversarial example by updating the direction of weight. By adding the cosine angle regularization which bounded in $[-1,1]$, the vanishing of the gradient caused by the increase of weight can be alleviated.}                      
	\label{fig:defense}                                             
\end{figure*}

\section{Methods}
In this section, we first analyze the shortcoming of standard PGD attack, and then propose our Scale-Invariant adversarial attack (SI-PGD) and Scale-Invariant (SI) adversarial defense mechanism.
\subsection{Shortcoming of standard PGD attack}
\label{sec:shortcoming}
One of the most effective attacks known till date is the PGD attack, which starts with a random initialization and moves along the gradient direction to maximize cross-entropy (CE) loss based on the output logits of the model.

One common training objective for deep neural networks is the CE loss, which is defined as
\begin{equation}
\mathcal{L}_{\rm CE}(f(\bx),y) =-\log f(\bx)_y=-g(\bx)_y+\log (\sum_{i=1}^{K}e^{g(\bx)_i}),
\end{equation}
where $g(\bx)_y$ is the logits output of the network with respect to class $y$, and $f(\bx)_y$ is the probability (softmax on logits) of $\bx$ belonging to class $y$.
Then the adversarial examples generated by PGD attack with maximum $L_\infty$ norm perturbation $\epsilon$ can be described as
\begin{align}
\label{equ:PGD}
&\bx_0^\prime = \bx+\mathbb{U}(-\epsilon,+\epsilon), \\
\bx_{t+1}^\prime = {\rm Clip}_{\bx,\epsilon}&(\bx_t^\prime +\eta \cdot {\rm sign}(\nabla_\bx \mathcal{L}_{\rm CE}(f(\bx_t^\prime),y))), \\
\hat{\bx}^\prime =&\arg \max_{\bx^\prime \in \mathcal{B}_\epsilon(\bx)} \mathcal{L}_{\rm CE}(f(\bx^\prime),y),
\label{equ:PGD2}
\end{align}
where $\mathbb{U}(\cdot)$ is the uniform function, ${\rm Clip}_{\bx,\epsilon}(\cdot)$ projects the adversarial example to satisfy the $\epsilon$ constrain, $t$ is the number of step and $\eta$ is the step size.
$\mathcal{B}_\epsilon(\bm{x})=\{\bm{x}^\prime:||\bm{x}^\prime-\bm{x}||_\infty\leq\epsilon\}$ denotes the $L_\infty$-norm ball centered at $\bm{x}$ with radius $\epsilon$. 

Analyzing the definition of PGD attack, we have
\begin{equation}
\begin{split}
\nabla_\bx \mathcal{L}_{\rm CE}(f(\bx_t^\prime),y)&=(-1+f(\bx_t^\prime)_y) \nabla_\bx g(\bx_t^\prime)_y\\&+\sum_{i \neq y} f(\bx_t^\prime)_i \nabla_\bx g(\bx_t^\prime)_i.
\end{split}
\end{equation}
If $f(\bx_t^\prime)_y \approx 1$ and consequently $f(\bx_t^\prime)_i \approx 0$ for $i \neq y$, then $\nabla_\bx \mathcal{L}_{\rm CE}(f(\bx_t^\prime),y) \approx 0$ and finite arithmetic yields $\nabla_\bx \mathcal{L}_{\rm CE}(f(\bx_t^\prime),y) = 0$. This phenomenon of gradient vanishing will make PGD attack ineffective. Note that one can achieve $f(\bx_t^\prime)_y \approx 1$ with a scaled classifier $h= f(\alpha \cdot g)$ equivalent to $f$ (i.e., they make the same decision for every $\bx$) but rescaled by a constant $\alpha \textgreater 0$.
The detailed proof is as follows:

\textit{Proof 1.} We denote $\hat{g}(\bx) = \alpha \cdot (\bm{W}^\top \bm{z}+b) =\alpha \cdot g(\bx)$ as the output logits of the classifier $h$, where $g(\bx)$ is the logits of classifer $f$. Then we have 
\begin{equation}
\hat{f}(\bx) =\mathbb{S}(\hat{g}(\bx)),
\end{equation}
where $\mathbb{S}(\cdot)$ is the softmax function. Considering the binary-class case with classes $k_1$ and $k_2$ as a example, we denote $k_1$ as the true label of input $\bx$, i.e., $\hat{f}(\bx)_{k_1} \textgreater \hat{f}(\bx)_{k_2}$. If we want $\hat{f}(\bx)_{k_1} \approx 1$, i.e.,
\begin{equation}
\begin{split}
\hat{f}(\bx)_{k_1} &=\frac{\exp(\alpha g(\bx)_{k1})}{\exp(\alpha g(\bx)_{k_1})+\exp(\alpha g(\bx)_{k_2})}\\
&=\frac{1}{1+\exp(\alpha(g(\bx)_{k_2}-g(\bx)_{k_1}))} \approx1,
\end{split}
\end{equation}
i.e.,
\begin{equation}
\exp(\alpha(g(\bx)_{k_2}-g(\bx)_{k_1})) \approx 0.
\label{equ:0}
\end{equation}
Denote 
\begin{equation}
\alpha(g(\bx)_{k_2}-g(\bx)_{k_1}) =\beta,
\end{equation}
where $\beta$ is a threshold that satisfies Equation (\ref{equ:0}) (e.g.,  $\beta \in [ -\infty, -10^5]$). Hereafter, we can get  
\begin{equation}
\alpha =\frac{\beta}{g(\bx)_{k_2}-g(\bx)_{k_1}}.
\end{equation}
Thus, we get a $\alpha$ proves that $f(\bx)_y \approx 1$ can be achieved by a classifier $h= f(\alpha \cdot g)$.
 
As we shown in Fig~\ref{fig:motivation}, when the logits rescaled by factor from $10^{-3}$ to $10^3$, the model accuracy attacked by PGD (blue line) changes obviously. 
This discovery exposes the instability of PGD based on the logits, which may influence the reported robustness of model evaluated by PGD.

\subsection{Scale-Invariant adversarial attack (SI-PGD)}
\label{sec:SI-PGD}
To mitigate this potential threat, we propose the Scale-Invariant adversarial attack (SI-PGD) based on the angle between the features in the penultimate layer and the weights in the softmax layer to guide the generation of adversaril examples.
Specifically, note that in Equation~(\ref{equ:f}) there is $\bm{W}^\top\bm{z}=(W_1^\top\bm{z},\cdots,W_K^\top\bm{z)}$, and $\forall k \in [K]$, the inner product $W_k^\top\bm{z}=\lVert W_k\rVert \lVert\bm{z}\rVert \cos(\theta_k)$, where we denote $\theta_k = \angle(W_k,\bm{z})$. Let $\cos \bm{\theta}=(\cos(\theta_1),\cdots,\cos(\theta_k))$, then the output prediction used in SI-PGD is described as
\begin{equation}
\widetilde{f}(\bx)=\mathbb{S}(\cos\bm{\theta}).
\end{equation}
The CE loss with $\widetilde{f}(\bx)$ is formulated as
\begin{equation}
\mathcal{L}_{\rm CE}(\widetilde{f}(\bx),y)=-\log \mathbb{S} (\cos \theta_y),
\end{equation}
where, $\forall \cos (\theta_y) \in [-1,1], \nabla_\bx\mathcal{L}_{\rm CE}(\widetilde{f}(\bx),y) \neq 0$. The proof is as follows:

\textit{Proof 2.} If $\nabla_\bx\mathcal{L}_{\rm CE}(\widetilde{f}(\bx),y) = 0$, i.e., $\widetilde{f}(\bx)_y \approx 1$.
Considering the binary-class case with classes $k_1$ and $k_2$ as a example, we denote $k_1$ as the true label of input $\bx$. i.e.,
\begin{equation}
\begin{split}
\widetilde{f}(\bx)_{k_1} & = \frac{\exp (\cos \theta_{k_1})}{ \exp(\cos (\theta_{k_1}) +\cos (\theta_{k_2}) )} \\
& = \frac{1}{1+\exp(\cos (\theta_{k_2})-\cos (\theta_{k_1})) }\approx 1,
\end{split}
\end{equation}
i.e.,
\begin{equation}
\exp(\cos (\theta_{k_2})-\cos (\theta_{k_1})) \approx 0.
\label{equ:1}
\end{equation}
Denote
\begin{equation}
\cos (\theta_{k_2})-\cos (\theta_{k_1}) = \beta,
\end{equation}
where $\beta$ is a threshold that satisfies Equation (\ref{equ:1}) (e.g.,  $\beta \in [ -\infty, -10^5]$). Hereafter, we can get  
\begin{equation}
\cos \theta_{k_1} =\cos\theta_{k_2} - \beta.
\label{equ:cos}
\end{equation}
Since $\cos \theta_{k_1} \in [-1,1]$, and $\cos \theta_{k_2} -\beta \in [-\infty, -10^5]$. Therefore, Equation (\ref{equ:cos}) does not hold. Thus, $\widetilde{f}(\bx)_y \approx 1$ as well as $\nabla_\bx\mathcal{L}_{\rm CE}(\widetilde{f}(\bx),y) = 0$ are impossible. 
Therefore, $\forall \cos (\theta_y) \in [-1,1], $ we can obtain that 
\begin{equation}
\nabla_\bx\mathcal{L}_{\rm CE}(\widetilde{f}(\bx),y) \neq 0.
\end{equation}

Therefore, there is no phenomenon of gradient vanishing for proposed SI-PGD.
Then similar to Equations (\ref{equ:PGD}) and (\ref{equ:PGD2}), we can obtain the proposed Scale-Invariant adversarial attack (SI-PGD) as
\begin{align}
\label{equ:SI-PGD}
\bx_{t+1}^\prime = {\rm Clip}_{\bx,\epsilon}&(\bx_t^\prime +\eta \cdot {\rm sign}(\nabla_\bx \mathcal{L}_{\rm CE}(\widetilde{f}(\bx_t^\prime),y))) \nonumber\\
={\rm Clip}_{\bx,\epsilon}(\bx_t^\prime &+\eta \cdot {\rm sign}(\nabla_\bx (-\log \mathbb{S}(s\cdot \cos\theta_y^\prime)_t)),
\end{align}
here $s \textgreater 0$ is a hyperparameter to improve the numcrical stability~\cite{wang2017normface}. 

The superiority of our SI-PGD can be confirmed from two aspects. First, the consine angle matrix is used to learn angularly discriminative representation of each input, which will guide SI-PGD to generate adversarial examples more effectively. Second, the angle of input will not change with the rescaling of the logits (i.e., the effect of SI-PGD will not be influenced by  classifier such as $h= f(\alpha \cdot g)$, where $\alpha \textgreater 0$). The same verifications can also be observed in Fig.~\ref{fig:motivation} yellow line, where our SI-PGD is more effective and stable than standard PGD attack.

\begin{table*}[t]
	\centering
	\caption{Formulations of the adversarial training frameworks without (\xmark) or with (\checkmark) SI. The differences between our enhanced versions (with SI) from the original versions (without SI) are colored red. The adversarial objective $\mathcal{L}_A$ constructs adversarial examples (the inner maximization problem), and the training objective $\mathcal{L}_T$ updates parameters (the outer minimization problem). ${\rm KL}(\cdot)$ is Kullback-Leibler (KL) divergence. }
	\setlength{\tabcolsep}{0.8mm}{
		\begin{tabular}{c|c|c|c}
			\toprule
			Methods & SI & Training objective  & Adversarial objective  \\
			\midrule
			\multirow{2}{*}{AT~\cite{madry2018towards}} & \xmark & $\mathcal{L}_{\rm CE}(f(\hat{\bx}^\prime),y)$ & $\mathcal{L}_{\rm CE}(f(\bx^\prime),y)$ \\
			\cmidrule(lr){2-4}
			& \checkmark & $\mathcal{L}_{\rm CE}(f(\hat{\bx}^\prime),y) + {\color{red} \beta  \mathcal{L}_{\rm CE}^m(\widetilde{f}(\hat{\bx}^\prime),y)}$  & $\mathcal{L}_{\rm CE}({\color{red}\widetilde{f}}(\bx^\prime),y)$\\
			\midrule
			\multirow{2}{*}{TRADES~\cite{zhang2019theoretically}} & \xmark & $\mathcal{L}_{\rm CE}(f({\bx}),y)+\lambda  \cdot {\rm KL}(f({\bx}),f(\hat{\bx}^\prime))$ & $ {\rm KL}(f({\bx}),f(\bx^\prime))$ \\
			\cmidrule(lr){2-4}
			& \checkmark & $\mathcal{L}_{\rm CE}(f({\bx}),y)+\lambda  \cdot {\rm KL}(f({\bx}),f(\hat{\bx}^\prime)) + {\color{red} \beta \mathcal{L}_{\rm CE}^m(\widetilde{f}(\hat{\bx}^\prime),y)}$  & ${\color{red}\mathcal{L}_{\rm CE}(\widetilde{f}(\bx^\prime),y)}$ \\
			\midrule
			\multirow{2}{*}{ALP~\cite{kannan2018adversarial}} & \xmark & $\alpha  \mathcal{L}_{\rm CE}(f({\bx}),y) +(1-\alpha)\mathcal{L}_{\rm CE}(f(\hat{\bx}^\prime),y) +\lambda \lVert \bm{W}^\top (\bm{z}-\hat{\bm{z}}^\prime)\rVert_2$ &  $\mathcal{L}_{\rm CE}(f(\bx^\prime),y)$  \\
			\cmidrule(lr){2-4}
			& \checkmark & $\alpha  \mathcal{L}_{\rm CE}(f({\bx}),y) +(1-\alpha)\mathcal{L}_{\rm CE}(f(\hat{\bx}^\prime),y) +\lambda \lVert \bm{W}^\top (\bm{z}-\hat{\bm{z}}^\prime)\rVert_2 +{\color{red} \beta \mathcal{L}_{\rm CE}^m(\widetilde{f}(\hat{\bx}^\prime),y)}$ & $\mathcal{L}_{\rm CE}({\color{red}\widetilde{f}}(\bx^\prime),y)$ \\
			\midrule
			\multirow{2}{*}{MART~\cite{wang2019improving}}  & \xmark & $\mathcal{L}_{\rm BCE}(f(\hat{\bx}^\prime),y) + \lambda \cdot {\rm KL}(f({\bx}),f(\hat{\bx}^\prime))\cdot(1-f(\bx)_y) $ & $\mathcal{L}_{\rm CE}(f(\bx^\prime),y)$ \\
			\cmidrule(lr){2-4}
			& \checkmark & $\mathcal{L}_{\rm BCE}(f(\hat{\bx}^\prime),y) + \lambda \cdot {\rm KL}(f({\bx}),f(\hat{\bx}^\prime))\cdot(1-f(\bx)_y) + {\color{red} \beta \mathcal{L}_{\rm CE}^m(\widetilde{f}(\hat{\bx}^\prime),y)}$  & $\mathcal{L}_{\rm CE}({\color{red}\widetilde{f}}(\bx^\prime),y)$\\
			\bottomrule		
	\end{tabular}}
	\label{tab:framework}
\end{table*}
\subsection{Scale-Invariant adversarial defense mechanism}

In this section, 
we discuss details on the proposed Scale-Invariant (SI) adversarial defense mechanism. 
As discussed in Section~\ref{sec:SI-PGD}, our proposed SI-PGD can obtain more effective and stable attack efficacy, so we utilize the adversarial examples generated by SI-PGD for adversarial training. Moreover, to obtain more robust defense, SI mechanism also applys cosine angle matrix as a regularization term in the outer minimization phase of the adversarial training, 
where $\widetilde{f}(\bx)$ is fed into the CE loss with a margin $m$ formulated as 
\begin{equation}
\mathcal{L}_{\rm CE}^m(\widetilde{f}(\bx),y)=-\log \mathbb{S}(s\cdot (\cos \theta_y-m)),
\end{equation}
where $m$ is used to learn angularly more discriminative features~\cite{wang2018cosface}.

As shown in Fig.~\ref{fig:defense}, we perform a toy example with a binary-class case to analyze our proposed SI mechanism during model outer minimization phase.
The decision boundary in CE loss is $(W_1-W_2)\bz +b_1-b_2=0$, where $W_i$ and $b_i$ are weights and bias in softmax layer, respectively. Assume that we have an adversarial example  $\hat{\bx}$ with original label $k_1$ and constrain $\lVert W_1 \rVert = \lVert W_2 \rVert$ and $b_1=b_2=0$, then the decision boundary becomes $\cos \theta_1 =\cos \theta_2$, as shown in Fig.~\ref{fig:defense1}.
Since the adversarial example is misclassified, $W_1$ will be updated towards both in norm (as shown in Fig.~\ref{fig:defense2}) and feature direction (as shown in Fig.~\ref{fig:defense3}), which may cause repetitive oscillation during training when fitting majority of adversarial examples.
Moreover, the unconstrained nature of the weight may cause the model to be more inclined to update the norm of the weight rather than the direction, and the gradually increasing weight of the example will cause the network to increase its output logits, which may occur the phenomenon of gradient vanishing as analyzed in Section~\ref{sec:shortcoming}.
Therefore, our cosine angle regularization is dedicated to assisting the model to update the weight in the direction of features, which alleviates the noneffective oscillation on the weight norm and the phenomenon of gradient vanishing due to the bounded value range $[-1,1]$.

To highlight our main contributions in terms of methodology, we summarize the proposed formulas of defenses in TABLE~\ref{tab:framework}. We mainly consider four representative adversarial training  frameworks including AT (adversarial training based on PGD)~\cite{madry2018towards}, TRADES~\cite{zhang2019theoretically}, ALP~\cite{kannan2018adversarial}, and MART~\cite{wang2019improving}. The differences between our enhanced versions (with SI) from the original versions (without SI) are colorized. Note that we apply the SI both on the adversarial objective for constructing adversarial examples (the inner maximization phase), and the training objective for updating parameters (the outer minimization phase).

\begin{table*}[t]
	\centering
	\caption{\textbf{The sensitivity of scale $s$.} Accuracy (\%) on CIFAR-10 with ResNet18. The training framework is TRADES-SI with different scale $s$ and margin $m=0.2$. The original accuracy of test images is shown in column ``Natural".}
	\centering
	\setlength{\tabcolsep}{4.5mm}{
		\begin{tabular}{c|c|c|c|c|c|c|c}
			\toprule
			Method & $m$ & $s$  & Natural & PGD & PGDCW & SI-PGD & AA \\
			\midrule
			TRADES & 0 & 0 & 84.15 & 55.23 & 52.67 & 52.24 & 50.06 \\
			\midrule
			\multirow{7}{*}{TRADES-\textbf{SI}} & 0.2 & 10 & 83.75 & 58.03 & 53.21 & 53.34 & 50.05 \\
			& \textbf{0.2} & \textbf{15} & \textbf{84.05} & \textbf{58.73} & \textbf{53.56} & \textbf{54.77} & \textbf{50.60} \\
			& 0.2 & 20 & 84.15 & 58.73 & 53.49 & 55.46 & 50.11 \\
			& 0.2 & 30 & 83.71 & 58.26 & 52.94 & 56.33 & 50.12 \\
			& 0.2 & 40 & 83.16 & 57.90 & 52.77 & 57.37 & 49.54 \\
			& 0.2 & 50 & 83.10 & 57.27 & 52.22 & 57.92 & 49.04 \\
			& 0.2 & 60 & 82.12 & 57.38 & 52.20 & 58.47 & 49.06 \\
			\bottomrule	
	\end{tabular}}
	\label{tab:s}
\end{table*}

\begin{table*}[t]
	\centering
	\caption{\textbf{The sensitivity of margin $m$.} Accuracy (\%) on CIFAR-10 with ResNet18. The training framework is TRADES-SI with different margin $m$ and scale $s=15$. The original accuracy of test images is shown in column ``Natural".}
	\centering
	\setlength{\tabcolsep}{4.5mm}{
		\begin{tabular}{c|c|c|c|c|c|c|c}
			\toprule
			Method & $s$ & $m$ & Natural & PGD & PGDCW & SI-PGD & AA \\
			\midrule
			TRADES & 0 & 0 & 84.15 & 55.23 & 52.67 & 52.24 & 50.06 \\
			\midrule
			\multirow{7}{*}{TRADES-\textbf{SI}} & 15 & 0.0 & 82.95& 58.01& 52.86& 54.73& 50.11\\
			& 15 & 0.1 & 83.10& 58.46& 52.98& 54.28&50.17 \\
			& \textbf{15} & \textbf{0.2} & \textbf{84.05} & \textbf{58.73} & \textbf{53.56} & \textbf{54.77} & \textbf{50.60} \\
			& 15 & 0.3 & 83.62& 57.82& 53.20& 53.89& 50.00\\
			& 15 & 0.4 & 83.49& 57.78&53.16&54.12& 50.17\\
			& 15 & 0.5 & 82.72& 57.58&53.19&54.15&50.13 \\
			& 15 & 0.6 & 83.37&57.35&52.71&53.53&49.36\\
			\bottomrule	
	\end{tabular}}
	\label{tab:m}
\end{table*}

\begin{table*}[t]
	\centering
	\caption{\textbf{The sensitivity of margin $\beta $.} Accuracy (\%) on CIFAR-10 with ResNet18. The training framework is TRADES-SI with different $\beta$. The margin $m=0.2$ and scale $s=15$. The original accuracy of test images is shown in column ``Natural".}
	\centering
	\setlength{\tabcolsep}{4.5mm}{
		\begin{tabular}{c|c|c|c|c|c|c}
			\toprule
			Method & $\beta$  & Natural & PGD & PGDCW & SI-PGD & AA \\
			\midrule
			TRADES & -  & 84.15 & 55.23 & 52.67 & 52.24 & 50.06 \\
			\midrule
			\multirow{7}{*}{TRADES-\textbf{SI}} & 0 & 81.58 & 59.59 & 52.67 & 52.95 & 49.59 \\
			& 0.1 & 83.04 & 58.53 & 53.05 & 53.95 & 50.05 \\
			& 0.2 & \textbf{84.05} & \textbf{58.73} & \textbf{53.56} & \textbf{54.77} & \textbf{50.60} \\
			&  0.3 & 84.36 & 57.96 & 53.27 & 54.50 & 50.34 \\
			&  0.4 & 84.82 & 58.12 & 53.51 & 54.55 & 50.55 \\
			&  0.5 & 84.15 & 57.82 & 53.11 & 54.23 & 50.22 \\
			&  0.6 & 84.29 & 57.95 & 53.27 & 54.26 & 50.08 \\
			\bottomrule	
	\end{tabular}}
	\label{tab:b}
\end{table*}

\begin{table*}[t]
	\centering
	\caption{\textbf{Attacks (CIFAR-10)}:Accuracy (\%) of various defenses (rows) against adversaries generated using different 20-step attacks (columns) under the $\ell_\infty$ bound with $\epsilon=8/255$. Architecture of each defense is described in the column ``Model". WideResNet is denoted by W (W-28-10 represents WideResNet-28-10), ResNet18 is denoted by RN-18. The original accuracy of each defense is described in the column ``Natural". $\dagger$ Additional data used for training.}
	\label{tab:attack}
	\centering
{
		\begin{tabular}{c|c|c|cccc|cccc}
			\toprule
			& \multirow{2}{*}{\textbf{Model}} & \multirow{2}{*}{\textbf{Natural}} & \multicolumn{4}{c|}{\textbf{Single run of the attack}}  &  \multicolumn{4}{c}{\textbf{5 random restarts}} \\
			& & & PGD & PGDCW &PGDLR &\textbf{SI-PGD}  &PGD & PGDCW &PGDLR &\textbf{SI-PGD}\\
			\midrule
			AT \cite{madry2018towards} & RN-50 & 85.46 & 56.64 & 54.95 & 57.47 & \textbf{54.65} & 56.10 & 54.47 & 56.95 & \textbf{54.41} \\
			TRADES~\cite{zhang2019theoretically} & W-34-10 & 84.92 & 57.16 & 55.45 & 57.15 & \textbf{55.27} & 56.84 & 55.20 & 56.79 & \textbf{55.02} \\
			Pre-training~\cite{hendrycks2019using} $\dagger$ & W-28-10 & 87.11 & 59.60 & 58.27 & 60.39 & \textbf{57.09} & 59.19 & 57.83 & 59.95 & \textbf{56.95} \\
			RST~\cite{carmon2019unlabeled} $\dagger$ & W-28-10 & 89.69 & 65.05 & 63.09 & 64.73 & \textbf{62.34} & 64.52 & 62.74 & 64.33 & \textbf{62.18} \\
			MART~\cite{wang2019improving} & RN-18 & 83.07 & 55.41 & 51.50 & 53.42 & \textbf{51.32} & 55.06 & 51.08 & 52.79 & \textbf{50.95} \\
			MART~\cite{wang2019improving} $\dagger$ & W-28-10 & 87.50 & 64.77 & 61.04 & 62.65 & \textbf{60.60} & 64.31 & 60.47 & 62.04 & \textbf{60.32} \\
			HYDRA~\cite{sehwag2020hydra} $\dagger$ & W-28-10 & 88.98 & 62.84 & 60.54 & 62.41 & \textbf{60.13} & 62.34 & 60.23 & 61.89 & \textbf{59.93} \\
			\bottomrule
	\end{tabular}}
\end{table*}

\begin{table*}[t]
	\centering
	\caption{\textbf{Multi-step defenses under white-box attack}: Accuracy (\%) of different defense mothods (rows) against various $L_\infty$ norm bound ($\epsilon=8/255$) white-box attacks (columns) on CIFAR-10 and CIFAR-100 datasets with ResNet18 and WideResNet. The original accuracy of each defense is described in the column ``Natural".}
	\label{tab:de_w_m}
	\centering
	\begin{tabular}{c|c|c|ccc|c | c|ccc|c}
		\toprule
		\multirow{2}{*}{Network}& \multirow{2}{*}{Method} & \multicolumn{5}{c|}{\textbf{CIFAR-10}}  & \multicolumn{5}{c}{\textbf{CIFAR-100}} \\
		\cmidrule(lr){3-7} \cmidrule(lr){8-12}  
		&& Natural & PGD & PGDCW & SI-PGD & AA  & Natural & PGD &PGDCW & SI-PGD & AA \\
		\midrule
		\multirow{8}{*}{ResNet18}&AT \cite{madry2018towards} & 85.57 & 54.12 & 53.15 & 52.47 & 48.49  & 60.48 & 31.11 & 29.44 & 29.37 & 25.57 \\
		&AT-\textbf{SI} & 83.03 & 55.76 & 52.93 & 52.89 & 48.15  & 57.52 & 31.94 & 28.72 & 29.68 & 25.30 \\
		\cmidrule(lr){2-12}
		&TRADES~\cite{zhang2019theoretically} & 84.15 & 55.23 & 52.67 & 52.24 & 50.06 & \textbf{62.05} & 31.39 & 27.10 & 26.68 & 25.04 \\
		&TRADES-\textbf{SI} & 84.05 & 58.73 & \textbf{53.56} & \textbf{54.77} & \textbf{50.60} & 60.17& 33.54 & 28.19 & 29.66 & 25.86 \\
		\cmidrule(lr){2-12}
		&ALP~\cite{kannan2018adversarial} & 84.25 & 56.11 & 53.26 & 53.06 & 50.30 & 58.19 & 31.16 & 27.89 & 27.61 & 25.47 \\
		&ALP-\textbf{SI} & \textbf{85.70} & 57.64 & 53.06 & 52.66 & 49.30 & 61.49 & 32.46 & 28.38 & 28.62 & 25.51 \\
		\cmidrule(lr){2-12}
		&MART~\cite{wang2019improving} & 80.99 & 57.11 & 52.38 & 52.27 & 48.71 & 54.58 & 33.37 & 28.68 & 28.54 & 26.10 \\
		&MART-\textbf{SI} &81.46 & \textbf{61.02}& 52.49 & 53.90 & 49.09 & 56.53 &\textbf{35.25} & \textbf{29.50} & \textbf{31.02} & \textbf{26.77} \\
		\midrule 
		\multirow{8}{*}{WideResNet-34-10}&AT \cite{madry2018towards} & 85.60 & 55.84 & 54.78 & 54.76 & 51.02  & 59.48 & 32.34 & 30.01 & 29.63 & 26.84 \\
		&AT-\textbf{SI} & 85.69 & 57.01 & 55.37 & 55.58 & 51.48  & 60.25 & 32.65 & 30.44 & 31.51 & 27.26 \\
		\cmidrule(lr){2-12}
		&TRADES~\cite{zhang2019theoretically} & 86.36 & 58.04 & 56.32 & 56.27 & 52.35 & \textbf{63.74} & 33.85 & 30.62& 30.16 & 27.24 \\
		&TRADES-\textbf{SI} & 86.53 & 60.14 & \textbf{56.40} & \textbf{57.51} & \textbf{53.22} & 62.39& 35.07 & 30.89 & 31.90 & 28.54 \\
		\cmidrule(lr){2-12}
		&ALP~\cite{kannan2018adversarial} & \textbf{87.31} & 57.54 & 55.73 & 55.42 & 52.49 & 63.18 & 32.13 & 29.54 & 29.18 & 27.02 \\
		&ALP-\textbf{SI} & 86.04 & 59.65 & 55.74 & 55.57 & 52.73 & 64.14 & 32.89 & 30.00 & 30.11 & 26.83 \\
		\cmidrule(lr){2-12}
		&MART~\cite{wang2019improving} & 84.41 & 59.93 & 55.55 & 55.74 & 52.23 & 55.28 & 34.59 & 30.92 & 30.84 & 28.21 \\
		&MART-\textbf{SI} &85.30 & \textbf{62.31}& 56.17& 57.07 & 52.01 & 58.90 &\textbf{36.47} & \textbf{30.98} & \textbf{32.92} & \textbf{28.36} \\
		\bottomrule 
	\end{tabular}
\end{table*}

\begin{table}[t]
	\centering
	\caption{\textbf{Multi-step defenses under white-box attack on SVHN}: Accuracy (\%) of different defense methods (rows) against various $L_\infty$ norm bound ($\epsilon=8/255$) white-box attacks (columns) on SVHN with ResNet18. The original accuracy of each defense is described in the column ``Natural".}
	\label{tab:de_svhn}
	\centering
	{
		\begin{tabular}{c|c|ccc|c }
			\toprule
			& Natural & PGD & PGDCW & SI-PGD & AA  \\
			\midrule
			AT \cite{madry2018towards} & \textbf{92.91} & 57.99 & 56.20 & 55.39 & 49.75  \\
			AT-\textbf{SI} & 92.34 & 60.05 & 57.08 & 57.11 & 51.19   \\
			\midrule
			TRADES~\cite{zhang2019theoretically} & 90.42 & 61.28 & 57.39 & 56.90 & 50.82  \\
			TRADES-\textbf{SI} & 89.66 & 62.68 & 57.93 & \textbf{58.87} & \textbf{52.13} \\
			\midrule
			ALP~\cite{kannan2018adversarial} & 91.79 & 61.42 & 57.38 & 56.88 & 51.68  \\
			ALP-\textbf{SI} & 90.02 & 61.43 & 57.49 & 58.49 & 52.56 \\
			\midrule
			MART~\cite{wang2019improving} & 91.84 & 62.45 & 56.50 & 55.54 & 49.64 \\
			MART-\textbf{SI} &90.09 & \textbf{64.32}& \textbf{58.01}& 58.57 & 51.96  \\
			\bottomrule 
	\end{tabular}}
\end{table}

\section{Experiments}
In this section, we first introduce the experimental settings including the networks and adversarial attack settings we used in this study. 
Second, we choose the parameters $s$, $m$, and $\beta$ for our further experiments by ablation study.
Then, we present the experimental results of the proposed SI-PGD attack, followed by details on evaluation of the proposed defense mechanism SI.
Finally, we analyze the effectiveness of our method.

\subsection{Experiment setting}
\subsubsection{Networks}

\textbf{ResNet18 and PreActResNet18 setup~\cite{pang2021bag,gowal2020uncovering}.} For each defense method, we apply the SGD with momentum 0.9, weight decay $5 \times 10^{-4}$ and an initial learning rate 0.01, which is divided by 10 at the 75-th, 90-th, and 100-th epoch. The total epoch is 120. 

\textbf{WideResNet setup.} For each defense method, we apply the SGD with momentum 0.9, weight decay $7 \times 10^{-4}$ and an initial learning rate 0.1, which is divided by 10 at the 75-th. The total epoch is 90.
\subsubsection{Attack setting}
We compare proposed SI-PGD with three types of attack: PGD (standard PGD attack with cross-entropy loss based on the output logits of the model~\cite{madry2018towards}), PGDCW (PGD attack with maximum margin loss proposed in C\&W~\cite{carlini2017towards}), and PGDLR (PGD attack with difference of logits ratio loss (DLR)~\cite{croce2020reliable}). The maximum $L_\infty$ perturbation $\epsilon=8/255$ for all attacks with 20 steps. The step size is $\epsilon/4$. 

\subsection{Ablation study}
In this part, we explore the sensitivity of parameters $s$, $m$, and $\beta$. Different choices of the scale $s$, the margin $m$, and the hyperparameter $\beta$ in SI-PGD and SI defense mechanism lead to different trade-offs between the clean accuracy and the adversarial robustness of the trained models. TABLE~\ref{tab:s}, TABLE~\ref{tab:m}, and TABLE~\ref{tab:b} show the sensitivities of $s$, $m$, and $\beta$, respectively. We choose $s=15$, $m=0.2$, and $\beta =0.2$ in our experiments.

\subsection{Evaluation of the proposed SI-PGD attack}

The performance of various defenses against different attacks on CIFAR-10 is shown in TABLE~\ref{tab:attack}. We present results for both a single run of the attack (with a budget of 20 iterations), as well as the worst-case accuracy across 5 random restarts (with an effective budget of $5 \times 20$ iterations). Notably, SI-PGD consistently outperforms all other attacks across all defenses. Particularly under the most common comparison setting (single run of the PGD attack on CIFAR-10), The attack success rate of SI-PGD improved $\sim 2\%$ on AT~\cite{madry2018towards}, $\sim 3\%$ on RST~\cite{carmon2019unlabeled} and even $\sim4\%$ on MART~\cite{wang2019improving}. A similar trend of improvement can be also observed for the random restarts of the attacks.

\subsection{Evaluation of the proposed Scale-Invariant adversarial defense mechanism}
In this part, we embed our SI defense mechanism into both the popular multi-step and single-step adversarial defenses against white-box and black-box attacks to verify the superiority of our SI defense mechanism. As presented in TABLE~\ref{tab:framework}, we set $\beta=0.2$ for all defense methods, and $\lambda=6,3,6$ for TRADES, ALP, and MART, respectively. The attack used for training is ${\rm PGD}$ with random start. The maximun perturbation $\epsilon=8/255$ and the number of step is 10 with step size $\epsilon/4$. The mini-batch size is 128.

\subsubsection{Performance under white-box attack}

\begin{table}
	\centering
	\caption{\textbf{Single-step defenses under white-box attack}: Accuracy (\%) of different defense methods (rows) against various $L_\infty$ norm bound ($\epsilon=8/255$) white-box attacks (columns) on CIFAR-10 with different models RN-18 (ResNet18) and PARN-18 (PreActResNet18). Original accuracy of each defense is described in ``Natural".}
	\label{tab:de_w_s}
	\centering
	\setlength{\tabcolsep}{1.5mm}{
		\begin{tabular}{c|c|cccc}
			\toprule
			Model & Method & Natural & PGD & PGDCW & SI-PGD \\
			\midrule
			\multirow{4}{*}{RN-18} & FreeAT~\cite{shafahi2019adversarial} & \textbf{84.45} & 47.69 & 47.14 & 47.31 \\
			& FreeAT-\textbf{SI} & 84.22 & 48.37 & 46.97 & 46.56 \\
			\cmidrule(lr){2-6}
			& GAT~\cite{sriramanan2020guided} & 80.49 & 54.47 & 50.47 & 51.89 \\
			& GAT-\textbf{SI} & 83.42 & \textbf{56.01} & \textbf{51.83} & \textbf{52.13} \\
			\midrule
			\multirow{4}{*}{PARN-18} & RFGSM-AT~\cite{wong2019fast} & \textbf{83.80} & 49.41 & 48.92 & 48.85 \\
			& RFGSM-AT-\textbf{SI} & 82.85 & \textbf{50.75} & \textbf{49.69} & \textbf{49.42} \\
			\cmidrule(lr){2-6}
			& GradAlign~\cite{andriushchenko2020understanding} & 80.14 & 49.42 & 48.23 & 48.12 \\
			& GradAlign-\textbf{SI} & 82.58 & 50.51& 49.64 & 49.35 \\
			\bottomrule
	\end{tabular}}
\end{table}

\begin{table}[t]
	\centering
	\caption{\textbf{Defenses under black-box attack:} Accuracy (\%) of various defense methods under SPSA attack with different batch sizes. Average decision boundary (ADBD) and robust accuracy (Rob.Acc) of defense methods against black-box hard-label RayS attack. The dataset is CIFAR-10 with ResNet18 under $L_\infty$ norm bound $\epsilon=8/255$. }
	\label{tab:de_b}
	\centering
	\setlength{\tabcolsep}{2.8mm}{
		\begin{tabular}{c|ccc|cc}
			\toprule
			& \multicolumn{3}{c}{\textbf{SPSA}}  & \multicolumn{2}{c}{\textbf{RayS}} \\
			\cmidrule(lr){2-4} \cmidrule(lr){5-6}  
			& 128 & 256 & 512 & ADBD & Rob.Acc  \\
			\midrule
			AT \cite{madry2018towards} & 59.81 & 57.85 & 56.00 & 0.0370 & 53.91 \\
			AT-\textbf{SI} & 60.51 & 58.29 & 56.40 & 0.0373 & 54.43 \\
			\midrule
			TRADES~\cite{zhang2019theoretically} & 59.67 & 57.67 & 55.97 & 0.0393 & 54.95 \\
			TRADES-\textbf{SI} & 60.35 & 58.38 & \textbf{56.71} & \textbf{0.0394} & 54.25  \\
			\midrule
			ALP~\cite{kannan2018adversarial} & 60.48 & 58.27 & 56.55 & 0.0388 & 54.79\\
			ALP-\textbf{SI} & \textbf{61.02} & \textbf{58.62} & 56.65 & 0.0386 &  \textbf{55.39}\\
			\midrule
			MART~\cite{wang2019improving} & 58.16 & 56.20 & 54.91 & 0.0376 & 53.42 \\
			MART-\textbf{SI} &59.16 & 56.90 & 55.23 & 0.0380 & 53.88 \\
			\bottomrule 
	\end{tabular}}
\end{table}

\begin{figure*}[t]
	\centering      
	\subfigure[]{
		\begin{minipage}{0.35\columnwidth}
			\centering                                                          
			\includegraphics[width=1\columnwidth]{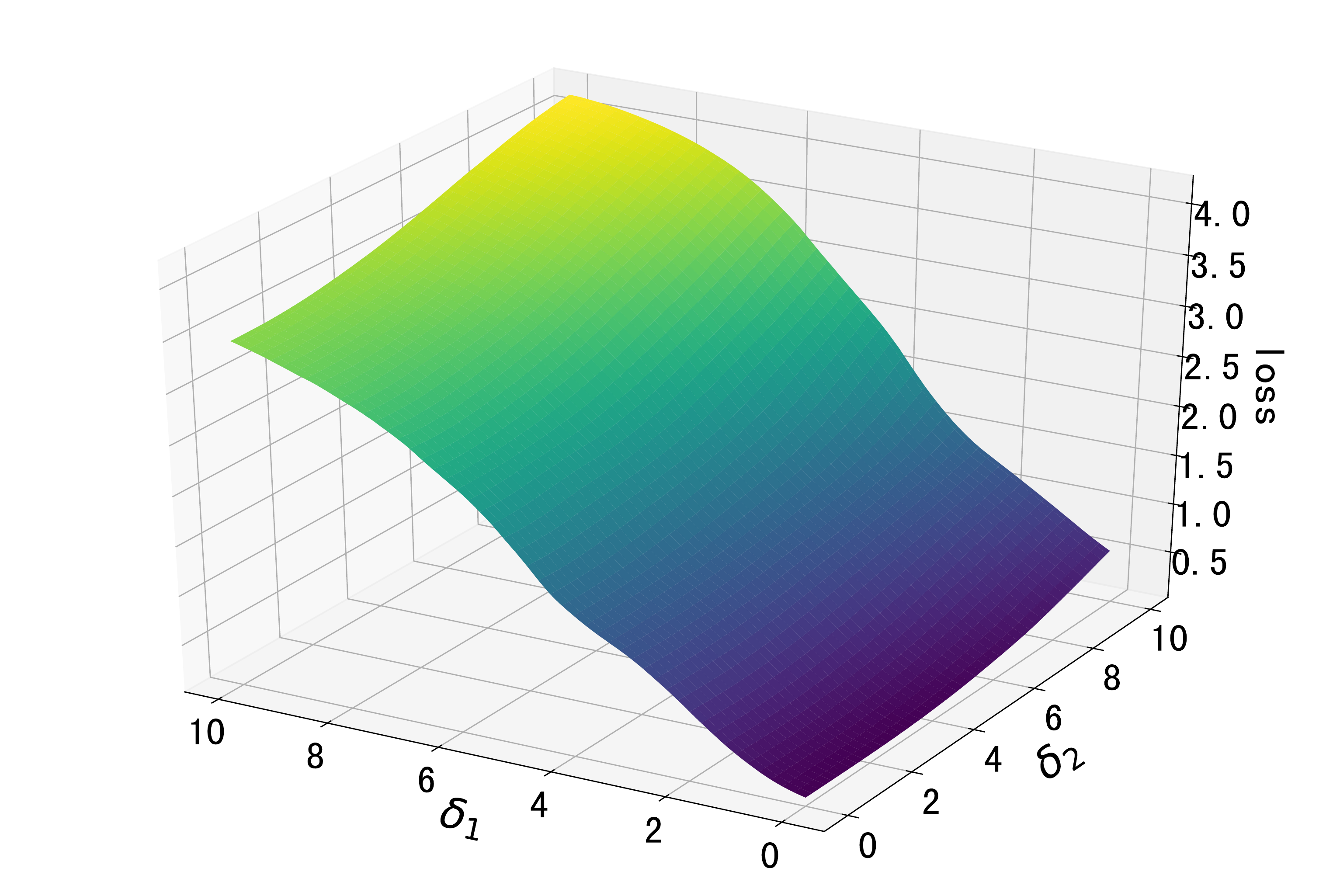}           
	\end{minipage}}
	\subfigure[]{
		\begin{minipage}{0.35\columnwidth}
			\centering                                                          
			\includegraphics[width=1\columnwidth]{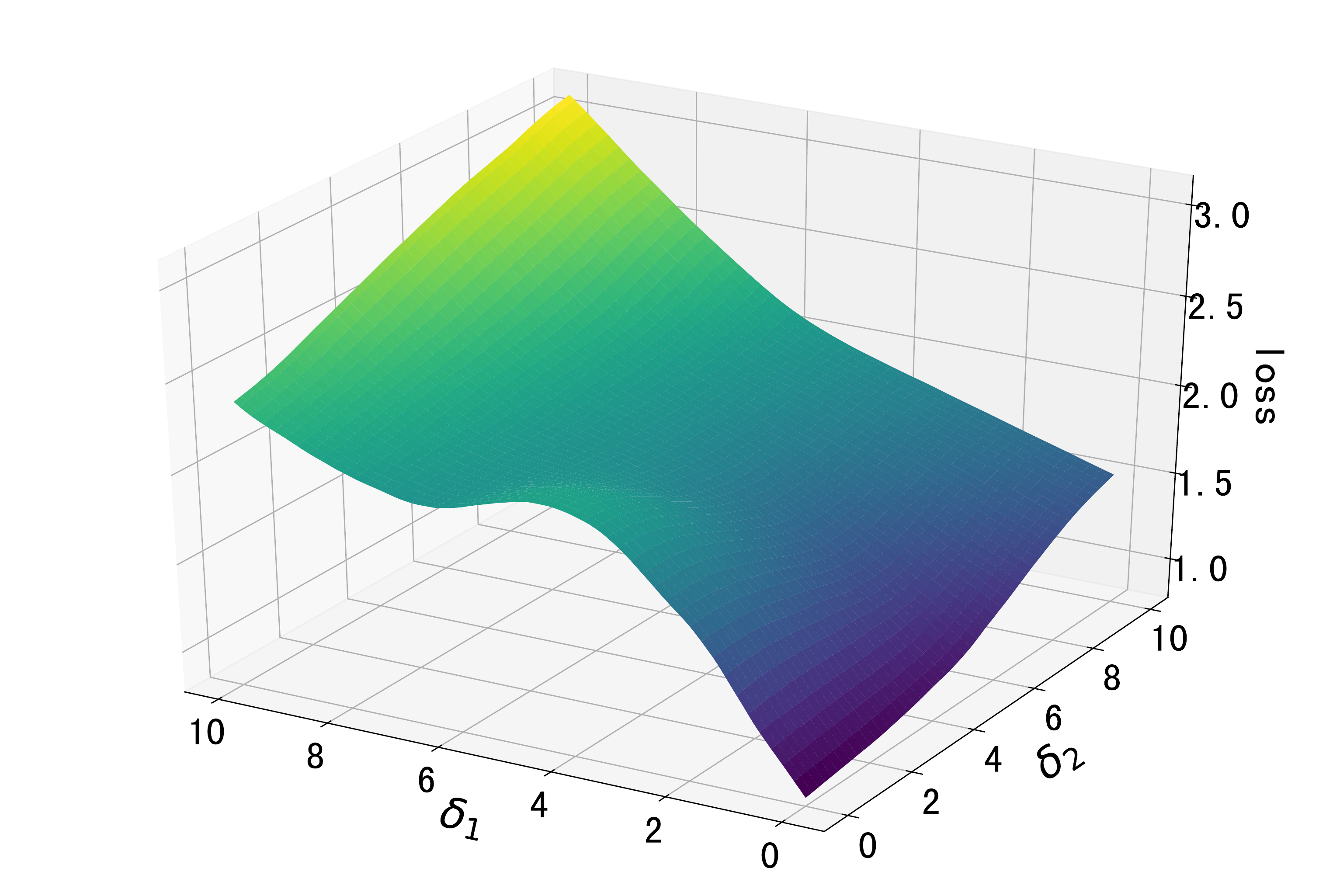}           
	\end{minipage}} 
	\subfigure[]{
		\begin{minipage}{0.35\columnwidth}
			\centering                                                          
			\includegraphics[width=1\columnwidth]{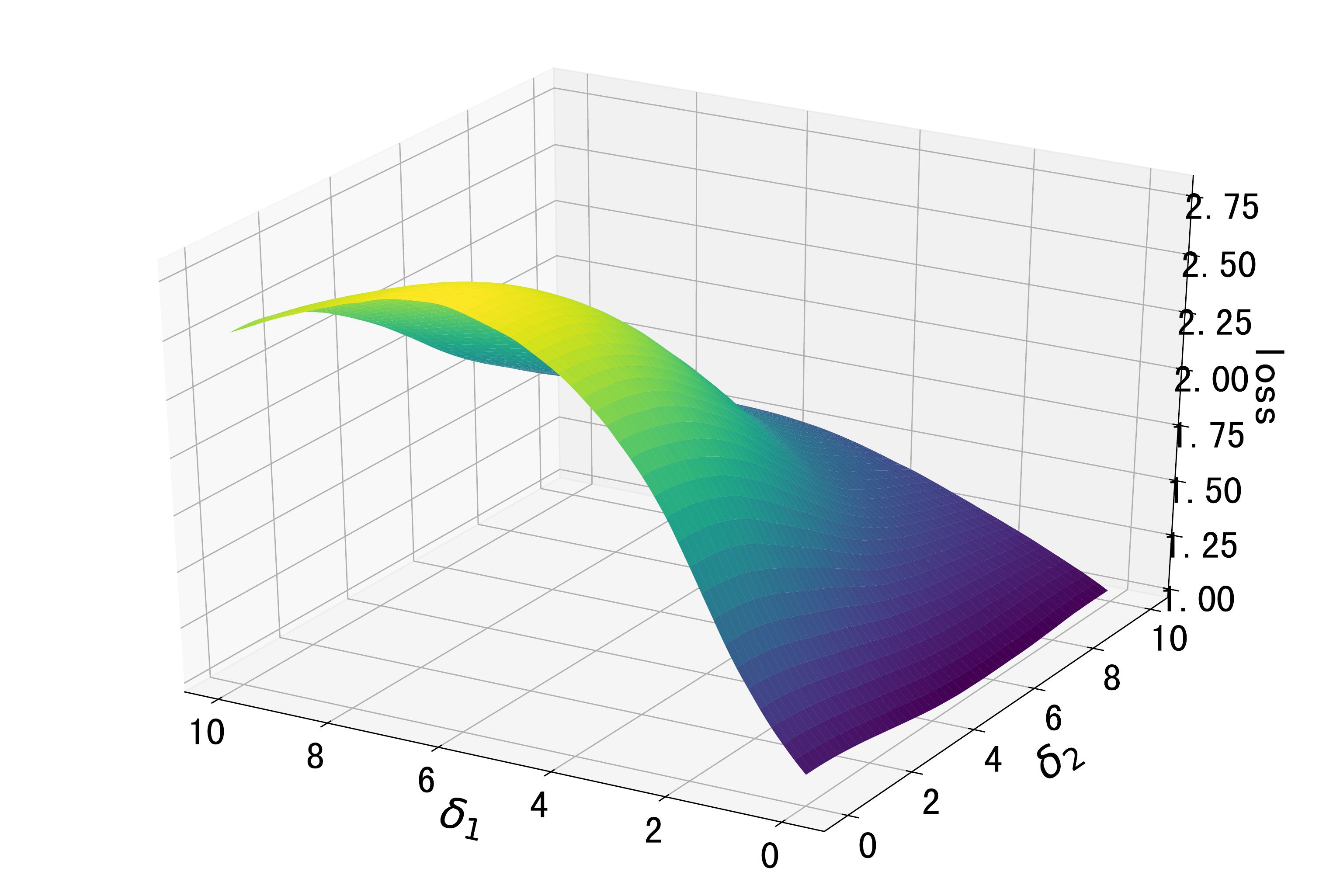}           
	\end{minipage}}
	\subfigure[]{
		\begin{minipage}{0.35\columnwidth}
			\centering                                                          
			\includegraphics[width=1\columnwidth]{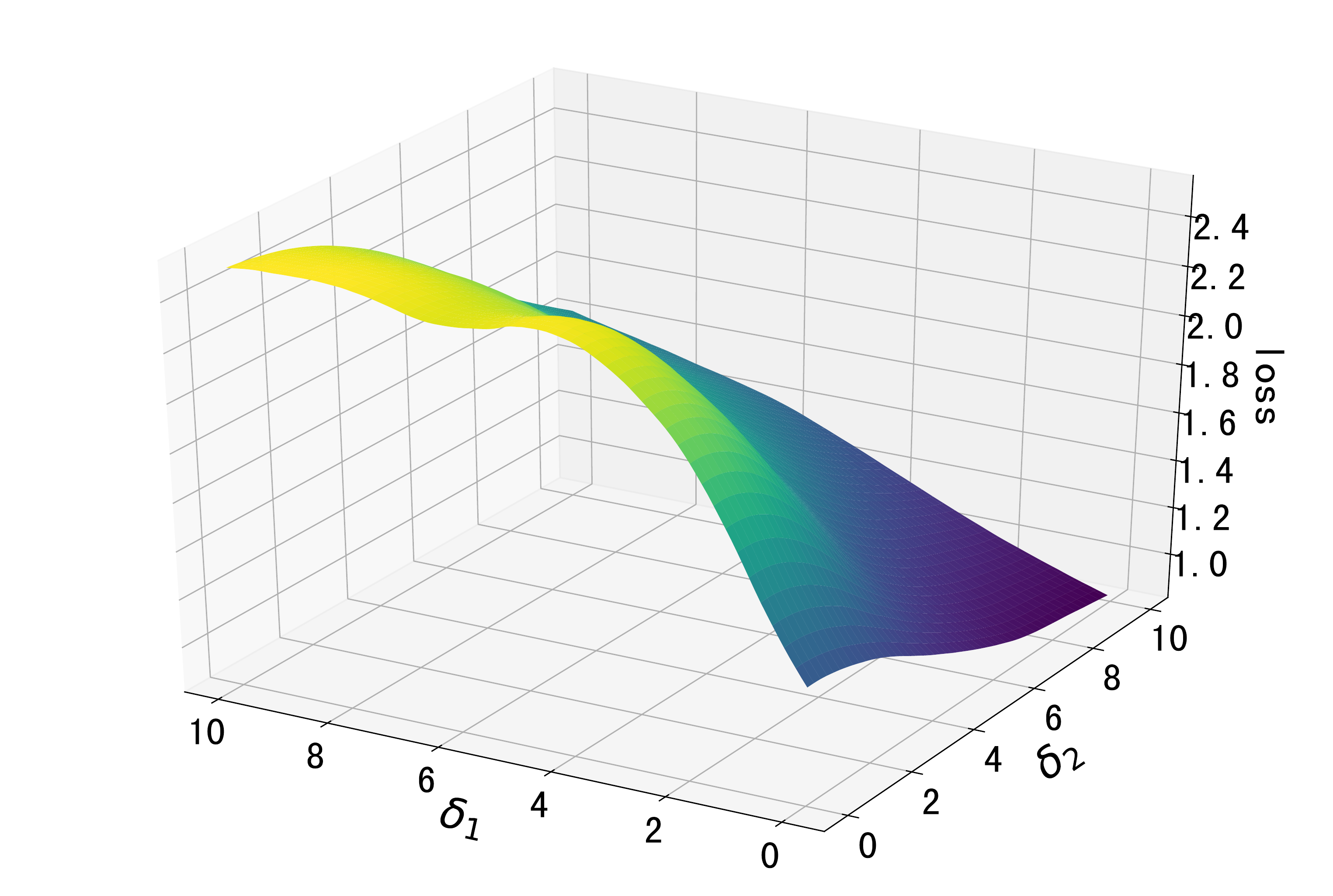}           
	\end{minipage}}
	\subfigure[]{
		\begin{minipage}{0.35\columnwidth}
			\centering                                                          
			\includegraphics[width=1\columnwidth]{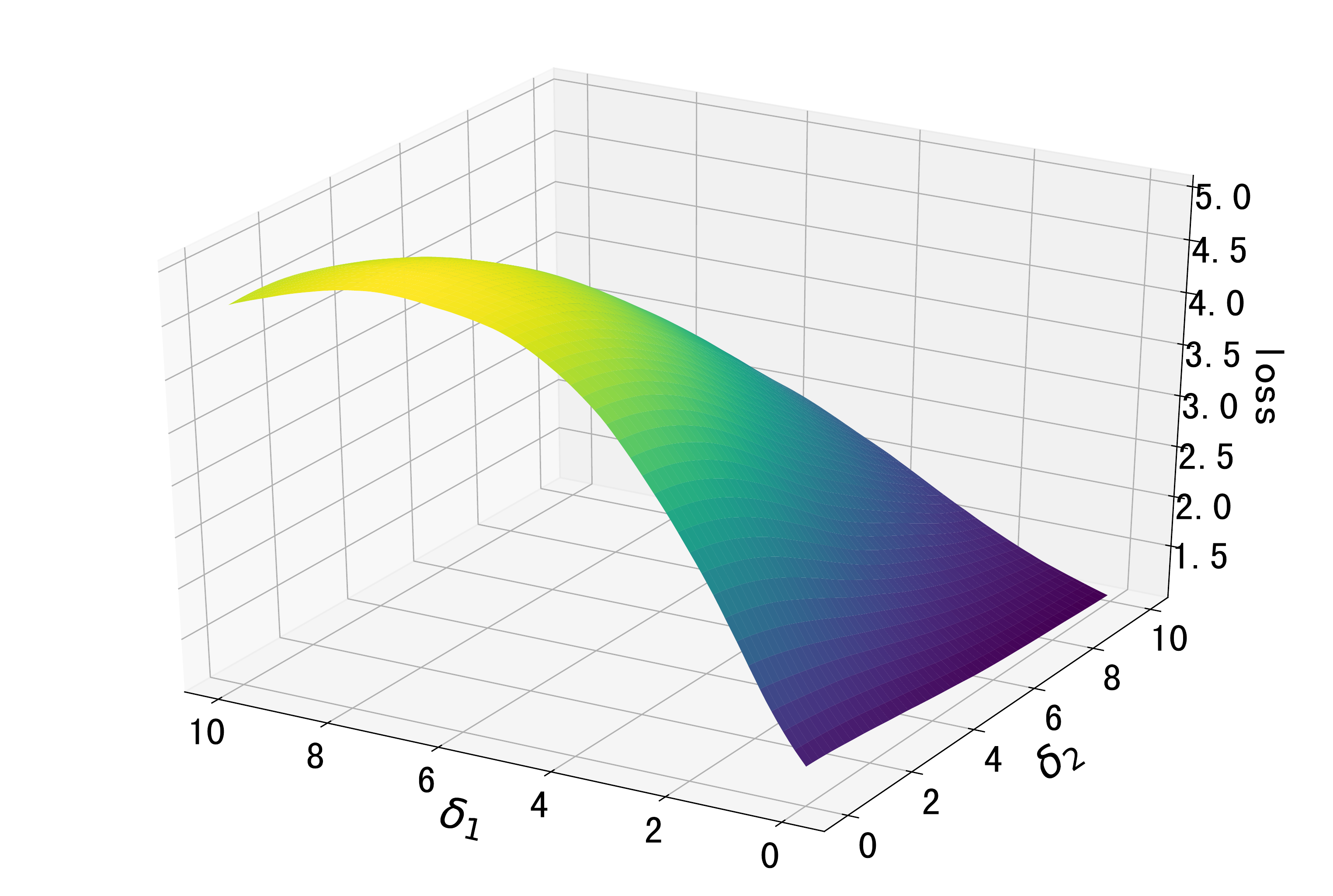}           
	\end{minipage}}
	\subfigure[]{
		\begin{minipage}{0.35\columnwidth}
			\centering                                                          
			\includegraphics[width=1\columnwidth]{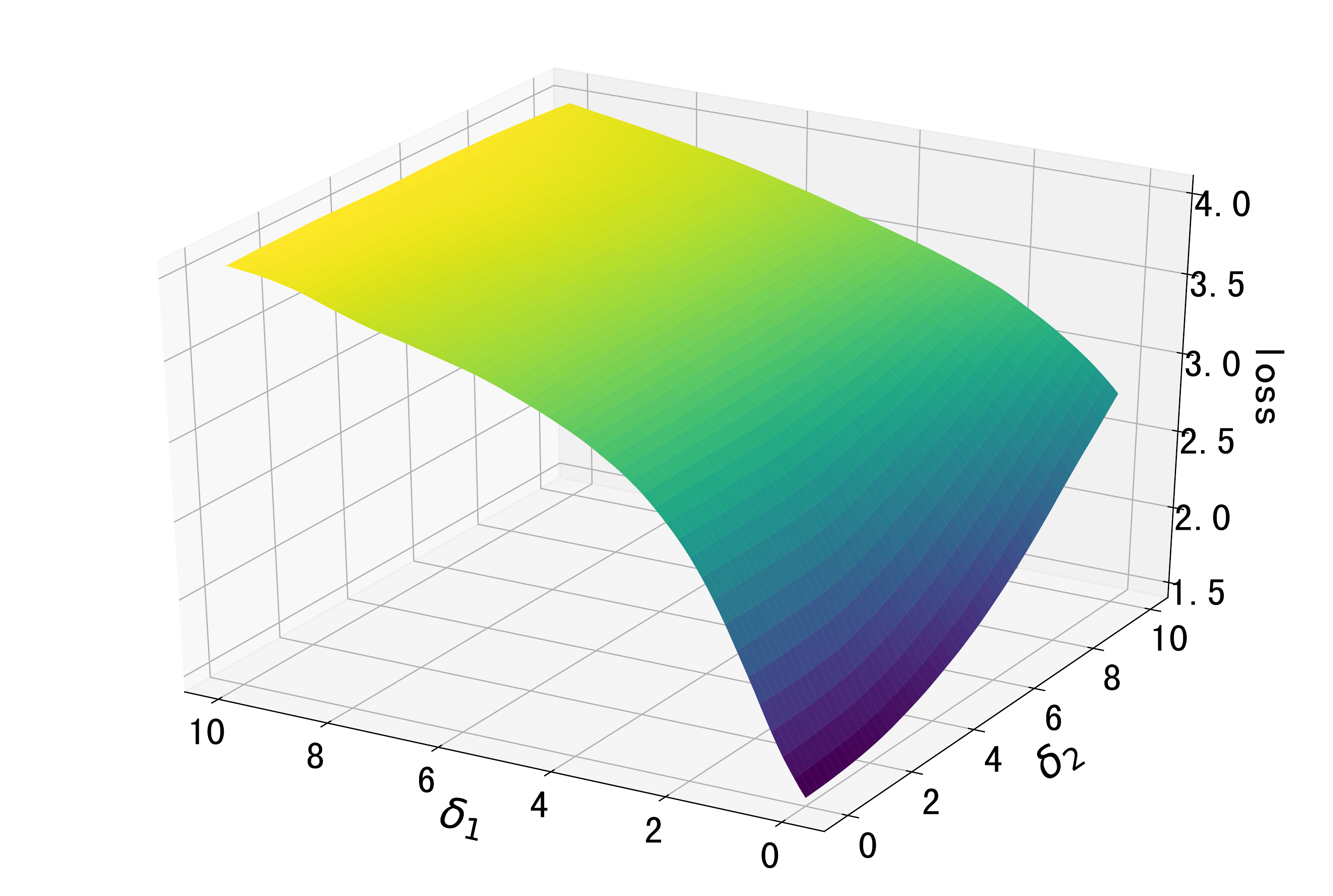}           
	\end{minipage}}
	\subfigure[]{
		\begin{minipage}{0.35\columnwidth}
			\centering                                                          
			\includegraphics[width=1\columnwidth]{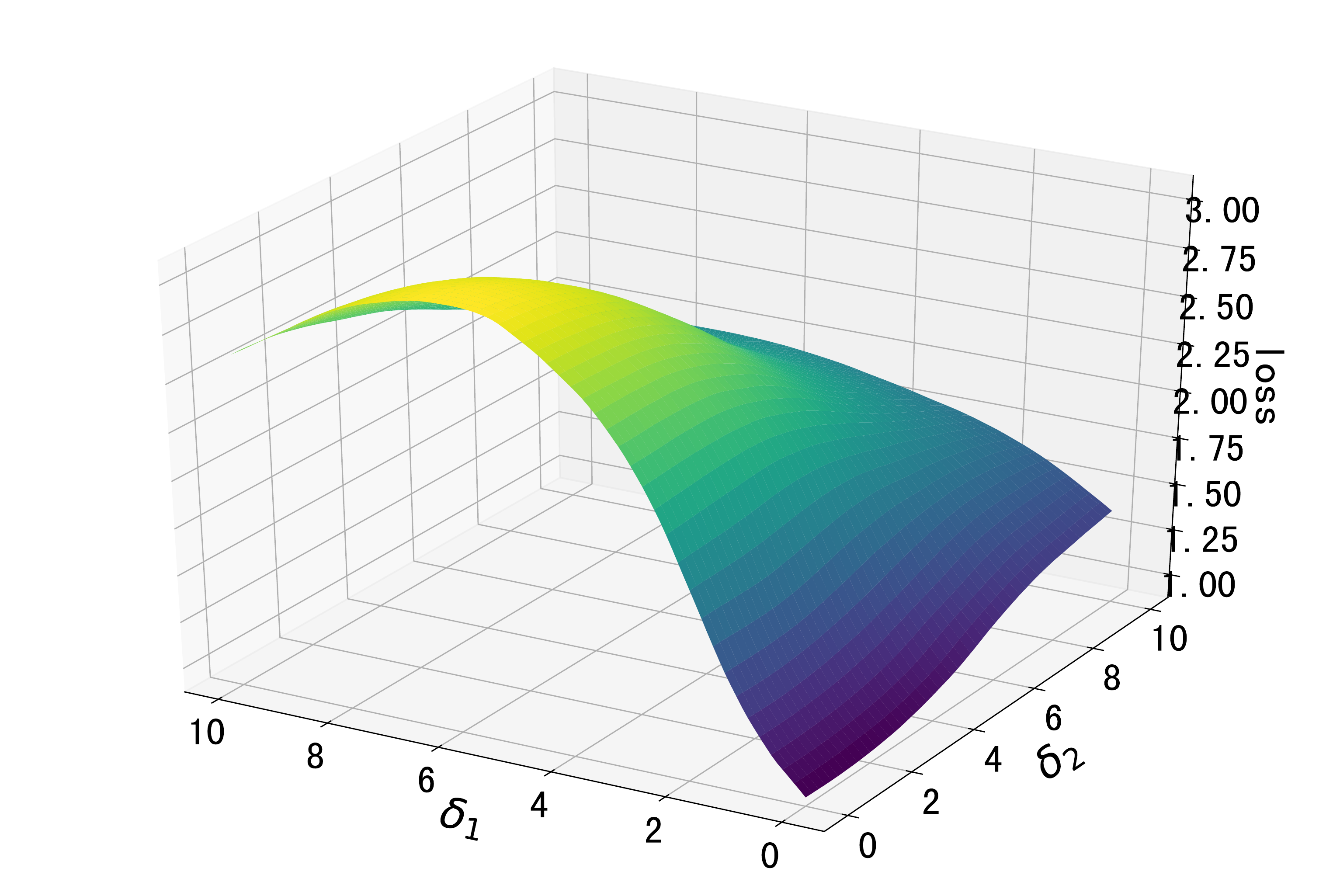}           
	\end{minipage}}
	\subfigure[]{
		\begin{minipage}{0.35\columnwidth}
			\centering                                                          
			\includegraphics[width=1\columnwidth]{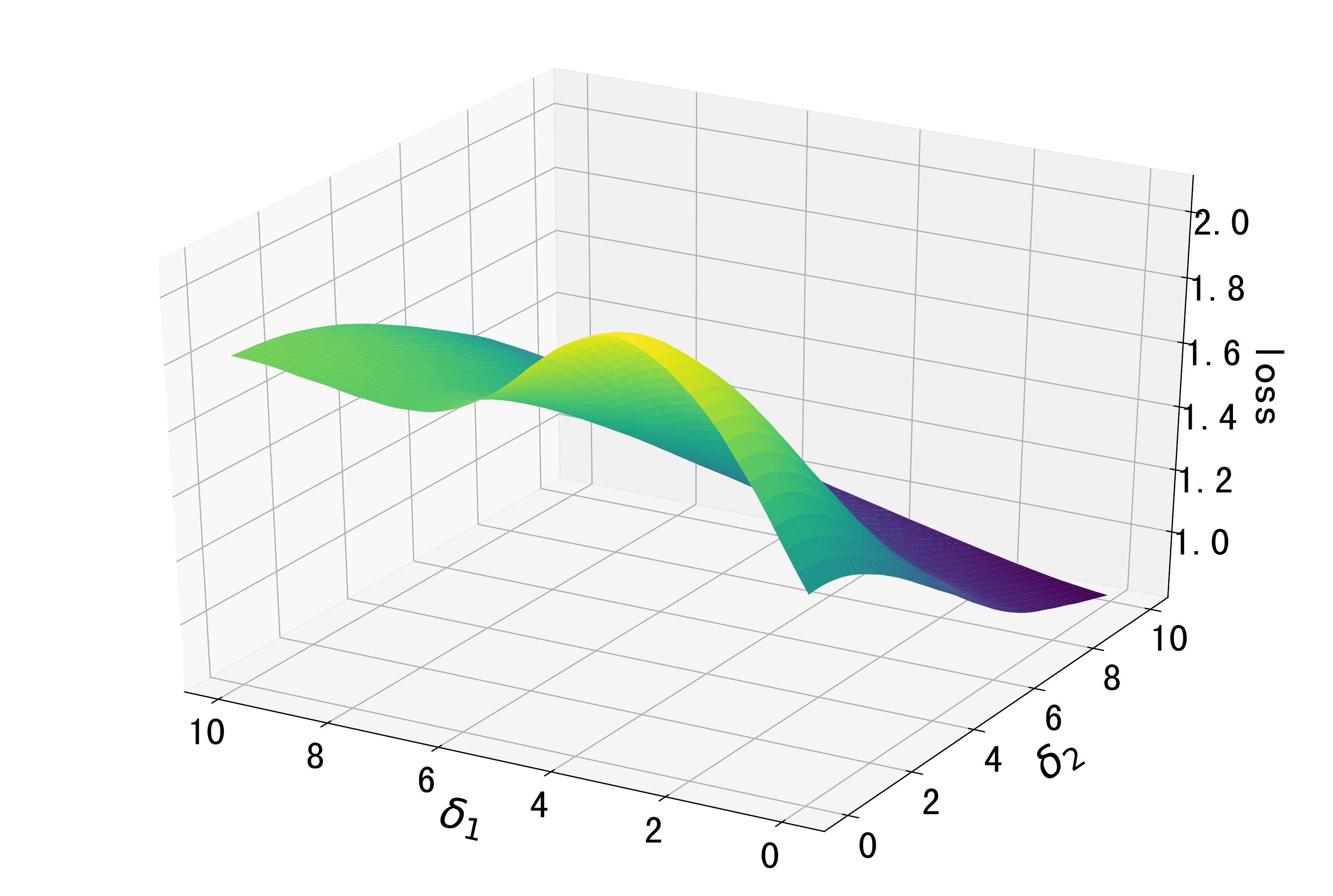}           
	\end{minipage}}
	\subfigure[]{
		\begin{minipage}{0.35\columnwidth}
			\centering                                                          
			\includegraphics[width=1\columnwidth]{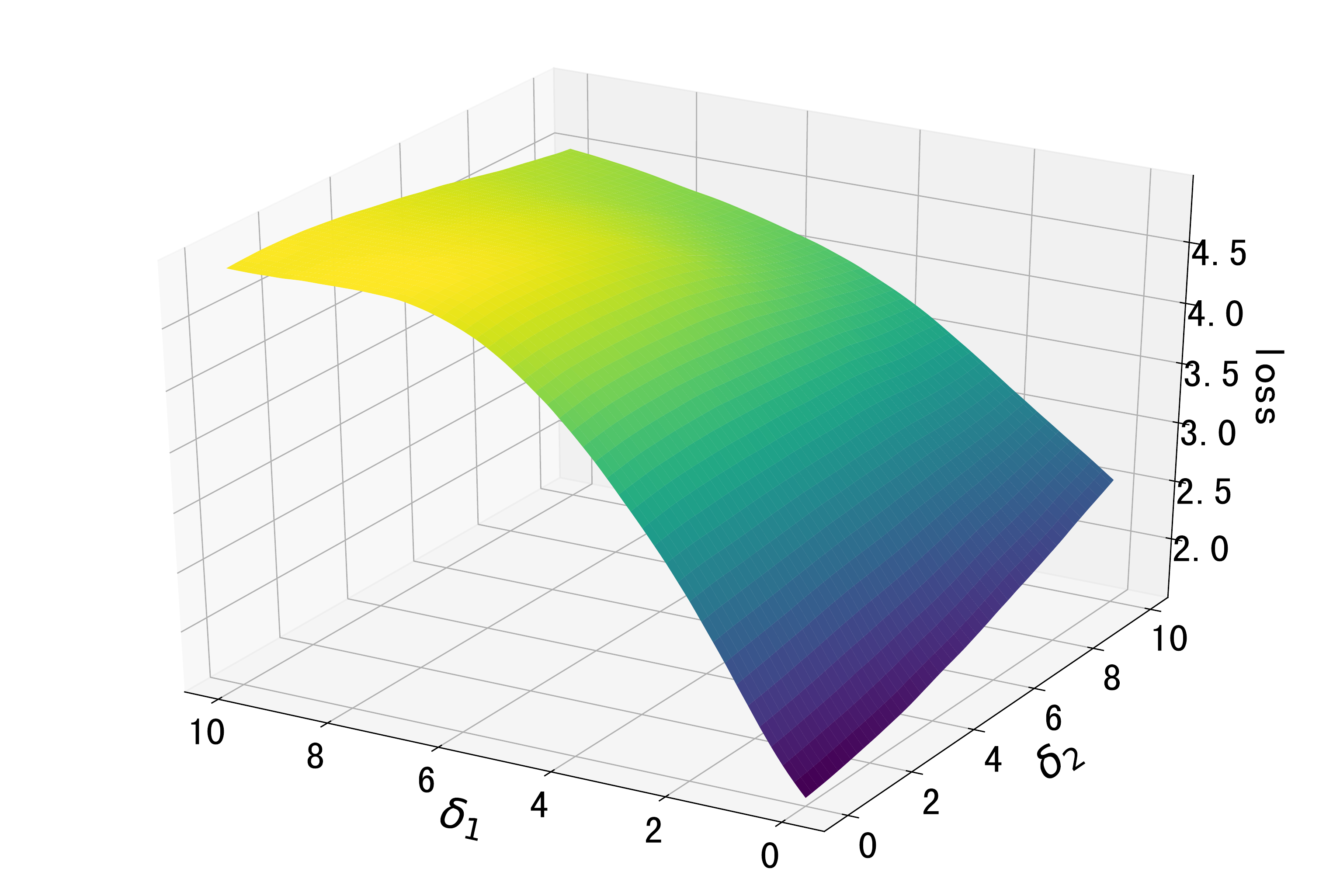}           
	\end{minipage}}
	\subfigure[]{
		\begin{minipage}{0.35\columnwidth}
			\centering                                                          
			\includegraphics[width=1\columnwidth]{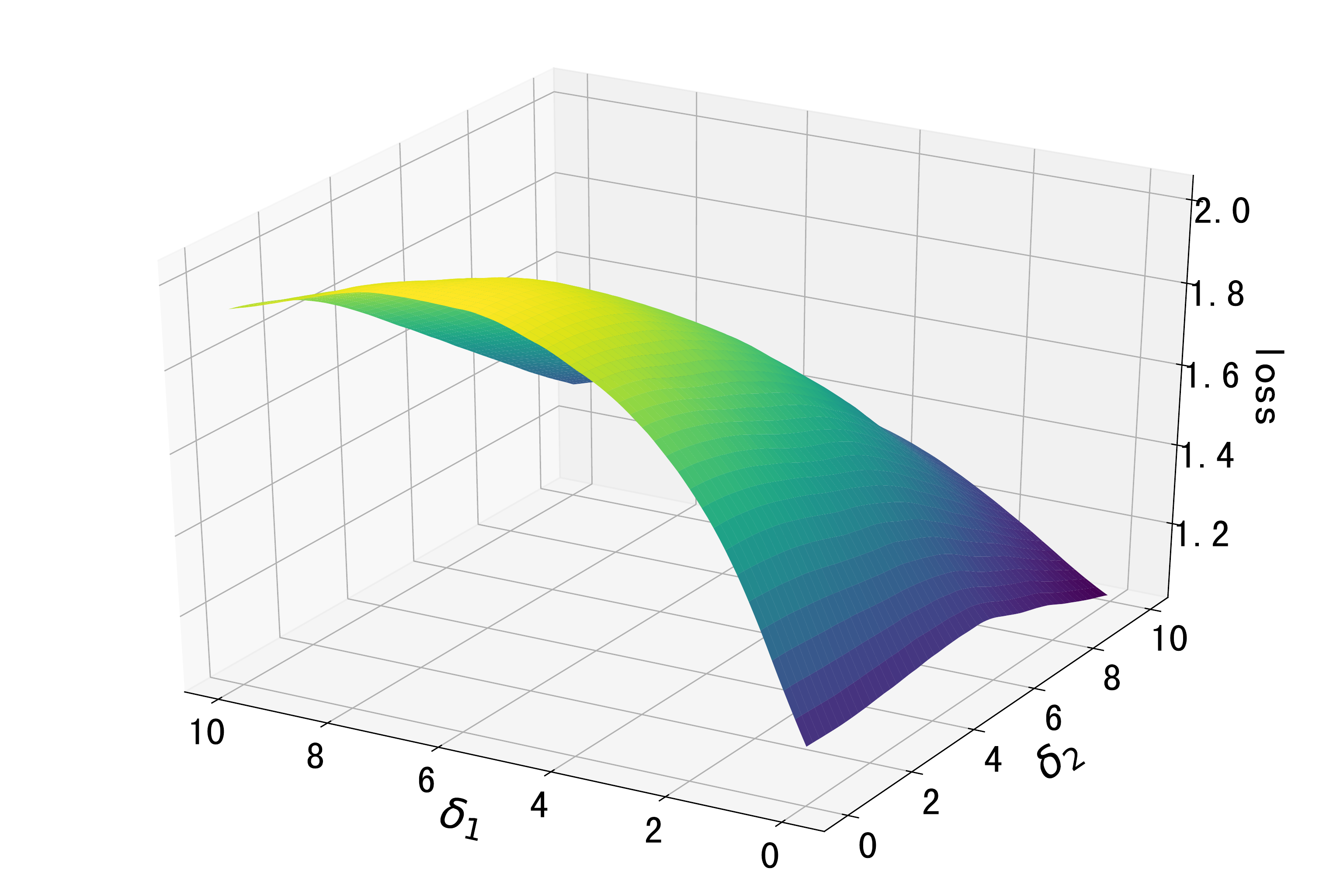}           
	\end{minipage}}
	\caption{\textbf{Example loss surfaces:} Plot of the loss surface of an AT-\textbf{SI} trained model with ResNet18 on CIFAR-10 perturbed images $\bx_i^\prime, i \in \{1,2,3,4,5,6,7,8,9,10\}$.}                      
	\label{fig:exloss}                                             
\end{figure*}

\begin{figure}[t]
	\centering                                                              
	\includegraphics[width=1\columnwidth]{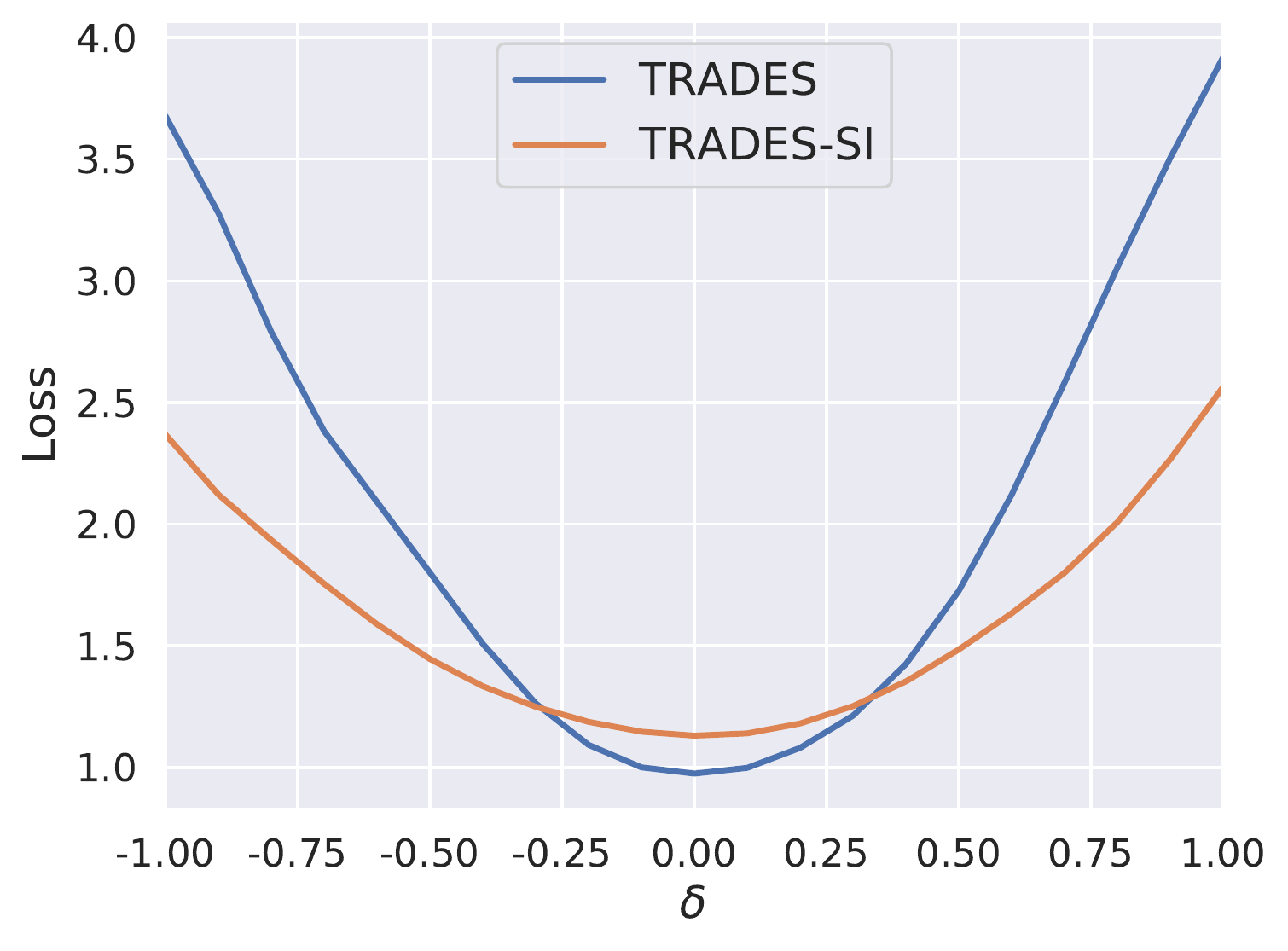}           
	\caption{\textbf{Weight loss surfaces} across different methods on CIFAR-10 with ResNet18.}                      
	\label{fig:wloss}                                             
\end{figure}


The white-box accuracy of existing multi-step defenses (i.e., AT~\cite{madry2018towards}, TARDES~\cite{zhang2019theoretically}, ALP~\cite{kannan2018adversarial}, and MART~\cite{wang2019improving}) on CIFAR-10, CIFAR-100, and SVHN are presented in TABLE~\ref{tab:de_w_m} and TABLE~\ref{tab:de_svhn}, respectively. The single-step defenses (i.e., FreeAT~\cite{shafahi2019adversarial}, GAT~\cite{sriramanan2020guided}, RFGSM-AT~\cite{wong2019fast}, and GradAlign~\cite{andriushchenko2020understanding}) on CIFAR-10 are presented in TABLE~\ref{tab:de_w_s}.

From TABLE~\ref{tab:de_w_m} we can observe, for multi-step defenses combined with our SI mechanism on CIFAR-10 with ResNet-18 and WideResNet against attacks, MART-SI shows the state-of-the-art defense accuracy when attacked by PGD, while TRADES-SI achieves the best defense accuracy attacked by PGDCW, AA and our proposed SI-PGD. The same improvement of defenses combined with SI can also be observed on CIFAR-100 dataset, where MART-SI achieves the best defense accuracy across all attacks. 

Moreover, from TABLE~\ref{tab:de_svhn} we can observe, for multi-step defenses combined with our SI mechanism on SVHN dataset against attacks, with the proposed SI defense mechanism, we obtain a 2.06\% defense accuracy increase on AT, 1.40\% increase on TRADES, and a 1.87\% incerease on MART against PGD attack.

We further analyze the impact of using the proposed SI defense mechanism in single-step defenses on CIFAR-10 dataset in TABLE~\ref{tab:de_w_s}. We observe significant robustness improvement in GAT-SI against all attacks compared with original GAT defense on ResNet18. Moreover, the proposed RFGSM-AT-SI outperforms the current state-of-the-art single-step defense on PreActResNet18 model by a significant margin. 
The all improvement in robustness significantly demonstrates the effectiveness of our proposed SI defense mechanism.

\subsubsection{Performance under black-box attack}

We perform black-box query-based SPSA attack~\cite{uesato2018adversarial} and hard-label RayS attack~\cite{chen2020rays} to fairly evaluate the effectiveness of our proposed SI. The results are reported in TABLE~\ref{tab:de_b}. 
For SPSA attack, to estimate the gradients, we set the batch size as 128, 256, and 512, the perturbation size as 0.001, and the learning rate as 0.01. We run SPSA attack for 100 iterations, and early-stop when we cause misclassification. As reported in TABLE~\ref{tab:de_b}, the defenses with SI obtain better robustness over original defenses in each setting. 

Furthermore, we record the Average Decision Boundary Distance (ADBD) and robust accuracy (Rob.Acc) against RayS in Table~\ref{tab:de_b}. On CIFAR-10 dataset, TRADES-SI achieves maximum ADBD and ALP-SI achieves best robust accuracy with ResNet18. 
As expected, these results show that embedding SI can generally provide promotion under the black-box threat models, and verify that SI defense mechanism reliably improves the robustness rather than causing gradient masking.


\subsubsection{The reliability of proposed scale-invariant (SI) adversarial defense mechanism}

Note that the robustness improvement of embedding SI is not caused by the so-called “obfuscated gradients”~\cite{athalye2018obfuscated}. 
This can be verified by four phenomenons: (1) Strong test attacks (e.g., PGDCW) have higher success rates (lower accuracies) than weak test attacks (e.g., PGD) as shown in TABLE~\ref{tab:attack}. (2) SI defense mechanism can defense AA effectively (i.e., TRADES-SI achieves 50.60\% accuracy when attacked by AA).
(3) Fig.~\ref{fig:exloss} shows the loss surfaces of our proposed AT-SI trained model on perturbed images of the form $\bx_i^\prime=\bx_i+\delta_1v+\delta_2r$, obtained by varying $\delta_1$ and $\delta_2$. We craft PGD attack with 10 steps perturbation as the normal direction $v$, and $r$ be a random direction, under the $L_\infty$ constraint of $\epsilon=8/255$. From Fig.~\ref{fig:exloss} we can observe, the loss surfaces of AT-SI trained model on perturbed images are smooth, which reveals the fact that the robustness improvement of our proposed SI is not caused by gradient obfuscation. 
(4) Following~\cite{wu2020adversarial}, we visualize the weight loss surface on TRADES-SI by plotting the adversarial loss change when moving the weight $\bm{W}$ along a random direction $d$ with magnitude $\delta$. As presented in Fig.~\ref{fig:wloss}, the weight loss surface of TRADES-SI is flatter (yellow line) than original TRADES (blue line), which indicates the effectiveness and stablity of our proposed SI defense mechanism.

The analysis above verify that our proposed SI mechanism improves adversarial robustness effectively and reliably rather than  gradient obfuscation or masking.

\subsection{Effectiveness analysis}

\begin{table}
	\centering
	\caption{\textbf{Comparision SI with HE}: Accuracy (\%) of different defense mechanism SI and HE (rows) against various $L_\infty$ norm bound ($\epsilon=8/255$) attacks (columns) on CIFAR-10, CIFAR-100, and SVHN, based on TRADES with ResNet18. }
	\label{tab:he}
	\centering
	{
		\begin{tabular}{c|c|ccc|c}
			\toprule
			Dataset& Method& PGD & PGDCW & SI-PGD & AA \\
			\midrule
			\multirow{3}{*}{CIFAR-10}&TRADES & 55.23 & 52.67  & 52.24 & 50.06 \\
			&TRADES-HE& \textbf{61.45} & 52.08  & 51.57 & 49.86 \\
			&TRADES-\textbf{SI} & 58.73 & \textbf{53.56}  & \textbf{54.77} & \textbf{50.60} \\
			\midrule
			\multirow{3}{*}{SVHN} & TRADES & 61.28 & 57.39 & 56.90 & 50.82 \\
			& TRADES-HE& \textbf{67.32} & 56.55 & 57.53 & 51.90 \\
			& TRADES-\textbf{SI} & 62.68 & \textbf{57.93} & \textbf{58.87} & \textbf{52.13} \\
			\midrule
			\multirow{3}{*}{CIFAR100} & TRADES & 31.39 & 27.10 & 26.68 & 25.04 \\
			& TRADES-HE& 31.70 & 25.51 & 29.34 & 23.93 \\
			& TRADES-\textbf{SI} & \textbf{33.54} & \textbf{28.19} & \textbf{29.66} & \textbf{25.86} \\
			\bottomrule
	\end{tabular}}
\end{table}

\subsubsection{Difference to hypersphere embedding (HE).}
The closest work to our SI defense mechanism is hypersphere embedding (HE)~\cite{pang2020boosting}, which normalizes the features in the penultimate layer and the weights in the softmax layer with an additive angular margin. HE replaces all logits in the loss function of the original defense method with cosine angle, while we use angle as a regularization to guide the training of the network. The accuracy of TRADES embedded with SI and HE under adversarial attacks (i.e., PGD, PGDCW, SI-PGD, and AA) 
are recorded in TABLE~\ref{tab:he}. From TABLE~\ref{tab:he} one can observe that, except for the PGD attack, the final robustness of TRADES-SI increases by a substantial amount than TRADES-HE.  
Note that the improvement on PGD performance of HE may be caused by the unstable property of PGD attack under rescaled logits (i.e., divided by feature norm and weight norm), as shown in Fig. \ref{fig:motivation}.
This also indicates that compared with HE using angle as the discriminant information, it is more effective to use angle as a regularization to assist model weight update. 
Only using angle without weight and feature norm as discriminant information may reduce the discriminative ability of the network.

\subsubsection{A close look at the angle learned by SI mechanism}
\begin{figure}[t]
	\centering                                                      
	\includegraphics[width=1.0\columnwidth]{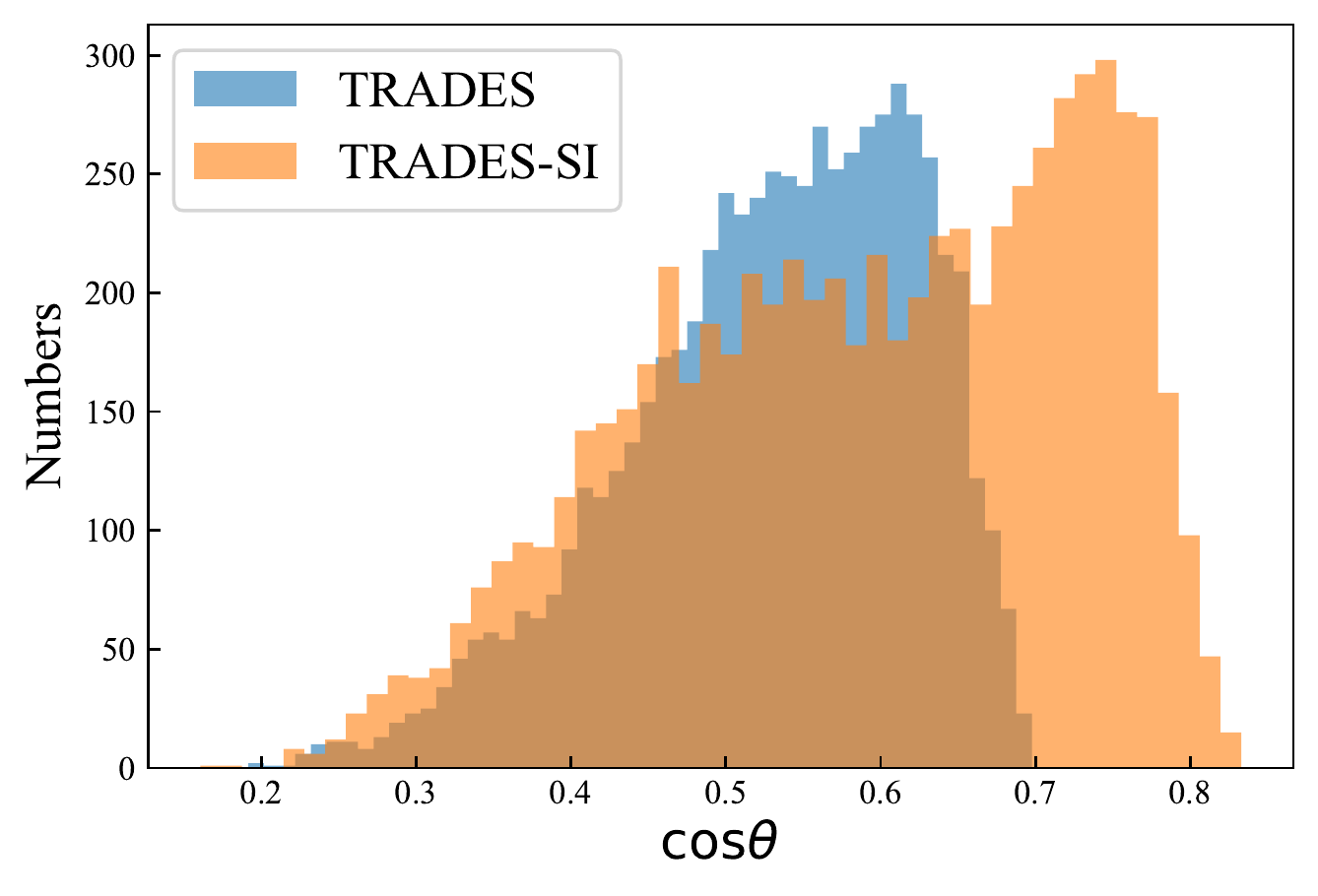}  
	\caption{The value distribution of $\cos \bm{\theta}$ on TRADES and TRADES-SI. The dataset is generated by PGD-10 on CIFAR-10 with ResNet18.}   
	\label{fig:weight} 
\end{figure}

In this part, we explore how the distribution of cosine angle changes when we apply SI defense mechanism on original defense method. 
We plot the histogram of $\cos \bm{\theta}$ values in different defense methods (i.e., TRADES and TRADES-SI) in Fig.~\ref{fig:weight}, and find that the $\cos \bm{\theta}$ of TRADES-SI (i.e., yellow bars) have larger values than TRADES (i.e., blue bars). This is because the regularization term added in the SI mechanism avoids the unlimited increasing of the network weight by increasing the cosine angle value, thereby avoiding the gradient vanishing and improving the robustness of the network. 

\section{Conclusion}

In this paper, we first reveal the unstability of the standard PGD attack, then we propose the Scale-Invariant adversarial attack (SI-PGD). We analyze the intriguing benefits induced by the angle between the model weight and the feature of the logits, and utilize the consine angle matrix of every input to guide the generation of adversarial examples. We demonstrate that our SI-PGD is consistently stronger than existing attacks across multiple defenses.
We further propose the Scale-Invariant (SI) adversarial defense mechanism, which is concise and adaptable to be embeded in the popular adversarial defenses. SI mechanism utilizes the proposed SI-PGD for both adversarial example generation and adversarial training regularization, thereby achieving a significant improvement in robustness over existing multi-step and single-step adversarial training methods.

\section{Acknowledgement}
This work was supported by the National Natural Science Foundation of China (Nos. 62136004, 61876082, 61732006), the National Key R\&D Program of China (Grant Nos. 2018YFC2001600, 2018YFC2001602), and also by the CAAI-Huawei MindSpore Open Fund.

\bibliographystyle{IEEEtran}
\bibliography{egbib}

%

%
%
%




\end{document}